\documentclass[runningheads]{llncs}

\usepackage[dvipsnames,table]{xcolor}
\usepackage[arxiv]{eccv}
\def\andname{}
\def\lastandname{\unskip,}
\def\pythoncodesize{\footnotesize}
\def\viscaptionsize{\scriptsize}
\def\visheight{6em}
\newcommand\visspacing{\vspace{5pt}\captionsetup[subfigure]{justification=centering,aboveskip=1pt,belowskip=2pt}}

\makeatletter
\renewcommand\paragraph{\@startsection{paragraph}{4}{\z@}%
                       {-12\p@ \@plus -4\p@ \@minus -4\p@}%
                       {-0.5em \@plus -0.22em \@minus -0.1em}%
                       {\normalfont\normalsize\bfseries\boldmath}}
\makeatother

\usepackage{chapterbib}

\usepackage{eccvabbrv}

\usepackage{graphicx}
\usepackage{booktabs}

\usepackage{makecell}
\usepackage{multirow}
\usepackage{soul}
\usepackage{afterpage}

\newcommand{\cellh}{\cellcolor{Goldenrod!35}}
\sethlcolor{Goldenrod!35}

\newcommand{\Sparo}{\textsc{Sparo}\xspace}

\usepackage{pythonhighlight}

\usepackage{amsmath,amsfonts,bm}

\def\1{\bm{1}}

\def\vtheta{{\bm{\theta}}}

\def\vh{{\bm{h}}}

\def\vm{{\bm{m}}}

\def\vq{{\bm{q}}}

\def\vx{{\bm{x}}}
\def\vy{{\bm{y}}}

\def\mH{{\bm{H}}}

\def\mK{{\bm{K}}}

\def\mW{{\bm{W}}}

\DeclareMathAlphabet{\mathsfit}{\encodingdefault}{\sfdefault}{m}{sl}
\SetMathAlphabet{\mathsfit}{bold}{\encodingdefault}{\sfdefault}{bx}{n}

\def\T{{\mathsf{T}}}

\newcommand{\R}{\mathbb{R}}

\newcommand{\softmax}{\mathrm{softmax}}
\newcommand{\cosim}{\mathrm{sim}}
\newcommand{\concat}{\mathrm{concat}}

\newcommand\norm[1]{\lVert#1\rVert}

\NewDocumentCommand{\pv}{m e{_} m}{%
  #1\IfValueT{#2}{_{#2}}^{(#3)}%
}

\DeclareMathOperator*{\E}{\mathbb{E}}

\usepackage[accsupp]{axessibility}  % Improves PDF readability for those with disabilities.

\usepackage[colorlinks,citecolor=eccvblue]{hyperref}

\usepackage{orcidlink}

\begin{document}

% ---------------------------------------------------------------

\title{\texorpdfstring{\includegraphics[height=18pt]{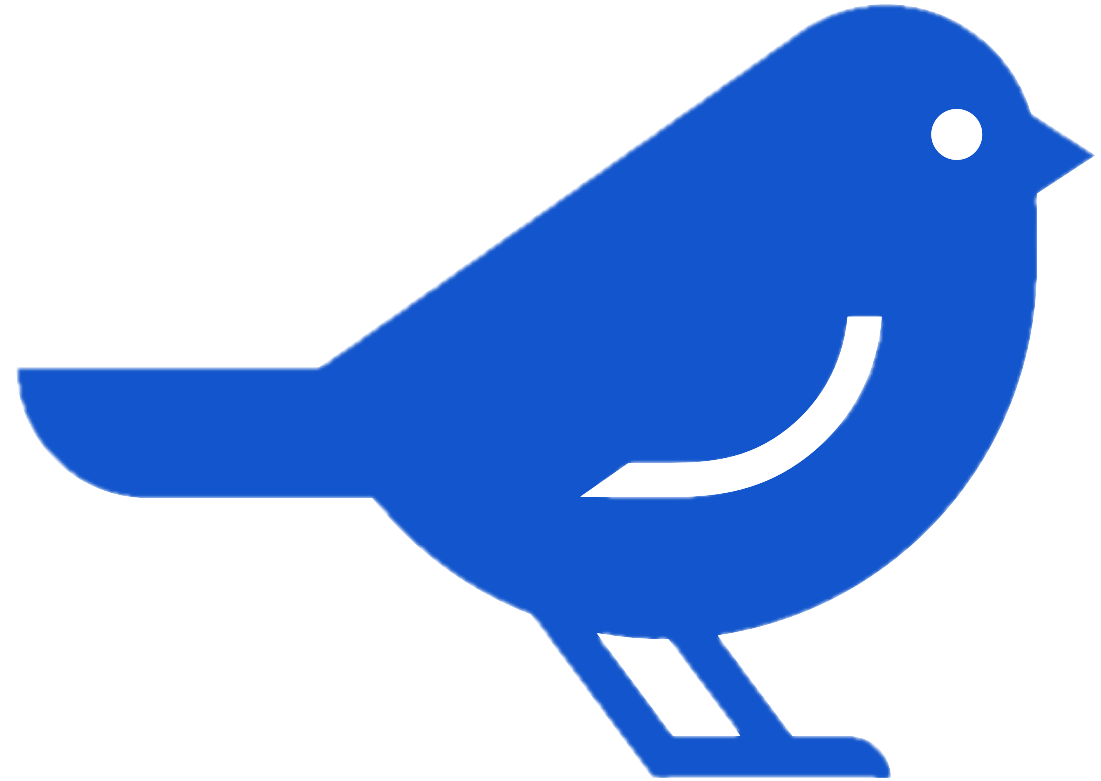}\,\textsc{\LARGE{Sparo}}}{SPARO}:\texorpdfstring{\,}{} Selective Attention for Robust and Compositional Transformer Encodings for Vision}

% Shorter title that can fit in one line of the running header
\titlerunning{\textsc{Sparo}: Selective Attention in Transformer Encodings for Vision}

% Author list. 
% Include the authors' ORCiD for the camera-ready version, if at all possible.
\author{Ankit~Vani\inst{1}\thanks{Correspondence at: \email{ankit.vani@umontreal.ca}.}\orcidlink{0009-0007-4781-9995} \and
Bac~Nguyen\inst{2}\orcidlink{0000-0001-9193-1908} \and
Samuel~Lavoie\inst{1}\orcidlink{0009-0005-8370-2415} \and
Ranjay~Krishna\inst{3}\orcidlink{0000-0001-8784-2531} \and
Aaron~Courville\inst{1,4}\orcidlink{0000-0001-6223-0301}}

% Abbreviated list of authors.
\authorrunning{A.~Vani et~al.}

% Institution list.
\institute{Mila, Université~de~Montréal \and
Sony~AI \and
University~of~Washington, Allen~Institute for Artificial Intelligence \and
CIFAR~AI~Chair}

\maketitle
\begin{abstract}

Selective attention helps us focus on task-relevant aspects in the constant flood of our sensory input. This constraint in our perception allows us to robustly generalize under distractions and to new compositions of perceivable concepts. Transformers employ a similar notion of attention in their architecture, but representation learning models with transformer backbones like CLIP and DINO often fail to demonstrate robustness and compositionality. We highlight a missing architectural prior: unlike human perception, transformer encodings do not separately attend over individual concepts. In response, we propose \Sparo, a read-out mechanism that partitions encodings into separately-attended slots, each produced by a single attention head. Using \Sparo with CLIP imparts an inductive bias that the vision and text modalities are different views of a shared compositional world with the same corresponding concepts. Using \Sparo, we demonstrate improvements on downstream recognition, robustness, retrieval, and compositionality benchmarks with CLIP (up to $+14\%$ for ImageNet, $+4\%$ for SugarCrepe), and on nearest neighbors and linear probe for ImageNet with DINO ($+3\%$ each). We also showcase a powerful ability to intervene and select individual \Sparo concepts to further improve downstream task performance (up from $+4\%$ to $+9\%$ for SugarCrepe) and use this ability to study the robustness of \Sparo's representation structure. Finally, we provide insights through ablation experiments and visualization of learned concepts.

\end{abstract}

\section{Introduction}
\label{sec:intro}

\emph{Selective attention} is an intrinsic property of human perception ~\cite{colby1999space,martinez2001putting,o2002attention}, enabling people to focus on task-relevant aspects of their surroundings while tuning out the rest~\cite{scott1962cognitive}. For instance, it allows a driver to focus on traffic signals, road signs, and other vehicles, or an individual searching for their friend in a crowd to attend to details like people's heights and clothing.
More broadly, it empowers us to comprehend the vast complexity of our world with limited mental resources \cite{boff1986handbook} by constraining our perception to one concept at a time.
When a task requires information for multiple concepts (\eg, looking at traffic signals and pedestrians), this constraint calls for sequentially attending to \emph{separate} salient aspects of our stimuli \cite{treisman1980feature}.
The cognitive bottleneck of selective attention, with the ability to separately represent concepts from our environment, enables human perception to remain insensitive to irrelevant distractions and generalize to new compositions of perceivable concepts.

In machine learning, the notion of selective attention is party emulated by the attention mechanism~\cite{bahdanau2014neural}, which is a core component of transformers~\cite{vaswani2017attention} such as the vision transformer (ViT)~\cite{dosovitskiy2020image}.
However, transformer encodings learnt using approaches like CLIP~\cite{radford2021learning} and DINO~\cite{caron2021emerging} still struggle with robustness \cite{zhai2019large,hendrycks2021natural,wang2019learning,hendrycks2021many} and compositional generalization \cite{ma2023crepe,hsieh2023sugarcrepe,thrush2022winoground,zhao2022vl,yuksekgonul2023when,ray2023cola}. For instance, despite CLIP emerging as the de~facto backbone for many vision tasks, it can perform close to chance when evaluated for compositionality~\cite{ma2023crepe}.
While the attention mechanism equips these models with the ability to emphasize relevant aspects of their intermediate representations, it does not endow them with the means to produce encodings for separately-attended concepts. Consequently, transformer-based representation learning models lack an important prior from human cognition: that perception of a complex input can be broken down into perception of its salient concepts.

This limitation deprives downstream tasks easy access to task-relevant aspects of the data's underlying structure. In a task like ImageNet classification, where most images contain one primary object of interest (\eg, ``dog''), information like relationships between objects and attributes of the background are irrelevant and can lead to overfitting if considered. However, these details can become crucial to attain high performance in other tasks like image retrieval (\eg, ``dog chasing a frisbee in a park''). Similar to the generalization benefits of selective attention in human perception, encodings that allow easy separation of attended concepts in the data can generalize to a diverse set of downstream tasks. Without additional constraints, however, the emergence of such compositional structure in encodings is unlikely, especially in the common setting of training on noisy internet data.

We propose \textbf{\Sparo} (\textbf{S}e\textbf{p}arate-head \textbf{a}ttention \textbf{r}ead-\textbf{o}ut)\footnote{Source code: \url{https://github.com/ankitkv/sparo-clip}.} as an improved read-out mechanism for transformer encoders in vision inspired by selective attention.
\Sparo replaces the last transformer block to provide a mechanism for partitioning encodings into \emph{slots} of separately-attended concepts.
We design each slot encoding to be a low-dimensional result of a \emph{single-head} attention operation, producing a bottleneck that encourages each slot to ``selectively attend.''
In this mechanism, producing multiple slots can be interpreted as selective attention over different concepts occurring in parallel.
When training CLIP, \Sparo imparts an inductive bias that both modalities are views onto a shared compositional world with the corresponding slots of both encoders representing its concepts. In contrast, standard CLIP merely embeds the encodings of both modalities in a shared vector space without imposing any additional structure.

Training \Sparo using CLIP and DINO exhibits improved generalization, robustness, and compositionality while retaining the same model size.
Our experiments show that using CLIP with \Sparo improves generalization for zero-shot recognition, robustness, retrieval, and compositionality (\eg, $+14\%$ for ImageNet, $+4\%$ for SugarCrepe, when trained with 15M examples; $+3\%$ each when trained with 400M examples). We also demonstrate improved linear probe performance of CLIP ($+10\%$ trained with 15M examples; $+2\%$ with 400M examples), as well as both linear probe and nearest neighbors performance of DINO ($+3\%$ each). We showcase how \Sparo's representational structure enables the ability to manually intervene and select relevant concepts for downstream tasks, leading to improved compositionality and generalization (up from $+4\%$ to $+9\%$ for SugarCrepe trained with 15M examples; $+3\%$ to $+6\%$ with 400M examples). We then study the robustness of this structure, validate the choice of separate-head attention through ablations, and provide insights into the effect of the number and dimensionality of concepts. Finally, we provide visualizations for some of the concepts \Sparo learns to attend to.

\section{Related work}
\label{sec:related}

We situate our work amongst other transformer read-out mechanisms and slot representation learning methods.

\paragraph{Transformer read-outs.}
For transformer~\cite{vaswani2017attention} decoders with causal masking, the last hidden state, corresponding to the end of sequence (\texttt{EOS}) input token, is used as the output representation. BERT~\cite{devlin2019bert} proposed inserting a special \texttt{CLS} token before input sequences for transformer encoders where information from all positions is gathered to produce the encoding. Using the \texttt{CLS} token output is also popular for Vision transformer (ViT)~\cite{dosovitskiy2020image,caron2021emerging,chen2021mocov3} backbones, along with alternatives such as global average pooling~\cite{chen2021mocov3,xu2022multi}. In the context of language, Multi-CLS BERT~\cite{chang2022multi} argues that an input text sequence can have multiple facets, and that samples can be similar along one facet and dissimilar along another. The authors aim to represent these facets by adding multiple \texttt{CLS} tokens to the input sequence. Using separate attention heads for each \Sparo slot can also encourage attending to different facets of the inputs, but without requiring expensive accumulations through each layer of the backbone. By keeping \Sparo independent of the backbone, we also allow for its easy potential placement on top of frozen pre-trained models without needing to train for added \texttt{CLS} tokens.

\Sparo is most related to the attentional pooler~\cite{lee2019set,yu2022coca} architecture, which uses multi-head attention with embedded learned queries to produce output encodings. However, \Sparo employs a critical structural constraint of not mixing the information from separate attention heads, a design choice we justify empirically in \cref{sec:exp:singlehead}. Slot attention~\cite{locatello2020object} also produces slot encodings through the attention mechanism, but rather than having the input positions compete for each slot, the slots compete for each input position. Slot attention also proposes an expensive iterative read-out mechanism where the role of each slot is determined by sampling from a per-sample Gaussian distribution, which can be limiting for large real-world datasets.

\paragraph{Slot representations.} Object-centric slot representations have been studied extensively in the literature~\cite{eslami2016attend,greff2019multi,burgess2019monet,zhang2019deep,engelcke2019genesis,locatello2020object,goyal2019recurrent,goyal2020factorizing,goyal2021neural,mansouri2024object,brady2023provably}. Often, these representations are expected to represent predefined objects and attributes, and evaluated in small-scale or synthetic settings. Recently, the use of slot attention on features produced by a learned encoder~\cite{seitzer2022bridging,qian2023semantics,aydemir2023self}, and utilizing unsupervised saliency masks for object discovery~\cite{zadaianchuk2022unsupervised} has enabled unsupervised learning of object-centric representations in larger scale settings. Related to these ideas, \Sparo attends to features produced by a transformer using a single attention head per slot, and each head is capable of learning saliency maps without supervision~\cite{caron2021emerging}. However, we do not constrain our slots to correspond to predefined types of objects in the data, nor do we impose any independence conditions between the slots. Self-supervised simplicial embeddings (SEM)~\cite{lavoie2023simplicial}, with only a soft-discrete slot structure constraint, enable improved generalization on downstream tasks. \Sparo replaces the separation of $\softmax$ application between SEM slots with a separation of underlying attention operations.

\section{Method}
\label{sec:method}

\begin{figure}[tb]
    \centering
    \includegraphics[width=\linewidth]{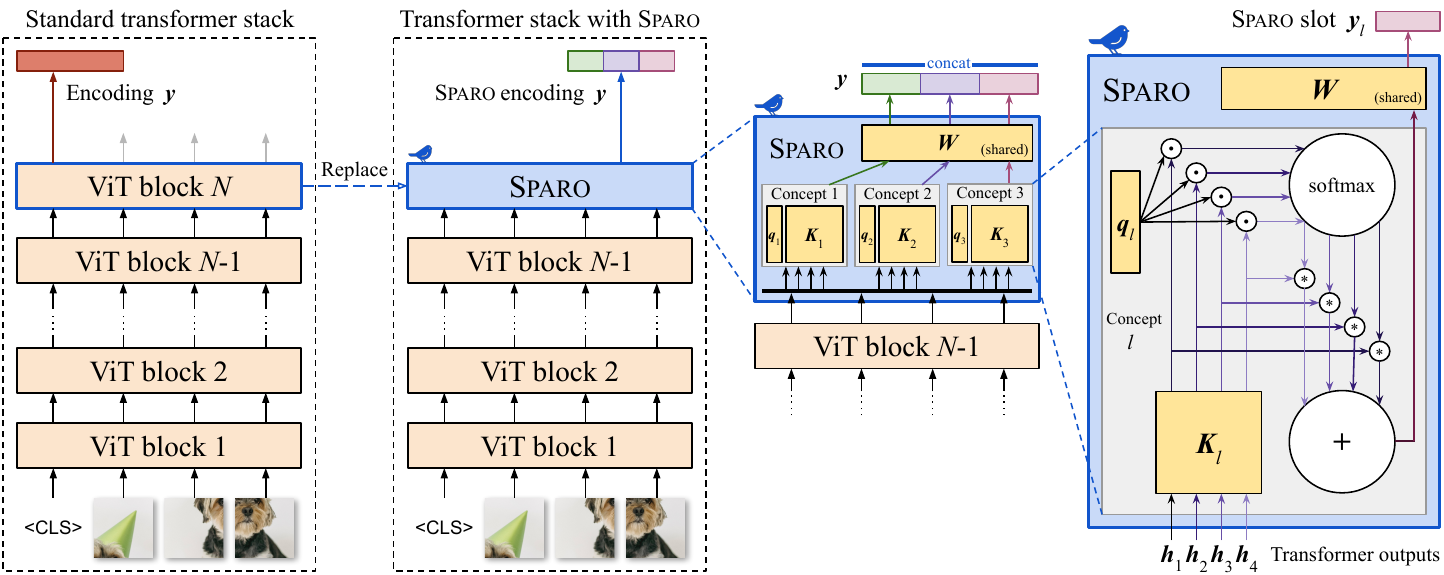}
    \caption{Illustration of \Sparo, a read-out mechanism that structures representations as collections of separately-attended concepts. Take a standard $N$-block transformer encoder (ViT here as an example), producing an encoding $\vy$ through extraction of its \texttt{CLS} token output. We can replace the $N$th transformer block with the \Sparo module (typically with equal or fewer parameters) to produce a \Sparo encoding $\vy$, which is a concatenation of $L$ \Sparo slots. Each \Sparo slot $\vy_l$ is produced through single-head attention over the backbone outputs using an embedded query $\vq_l$. The value projection is a composition of slot-specific key projection parameterized by $\mK_l$ and a slot-wise projection shared between all \Sparo concepts parameterized by $\mW$.}
    \label{fig:sparo}
\end{figure}

Having motivated our approach, we now detail the design of \Sparo. We start by introducing the relevant notation, then discuss the \Sparo module. Finally, we discuss the addition of \Sparo in CLIP and DINO, and the applicability of \Sparo beyond these settings.

\subsection{Notation}
\label{sec:notation}

We envision \Sparo as a special attention-based layer that modifies the top block of a transformer encoder~\cite{vaswani2017attention,radford2019language} backbone (including the ViT~\cite{dosovitskiy2020image}). Transformers take as input a sequence $\vx = \{\vx_1, \ldots, \vx_{n}\}$ of length $n$, and produce an output state per position $\{\vh_1, \ldots, \vh_{n}\}$, where $\vx_i \in \R^\text{input\,dim}$, $\vh_i \in \R^{d}$, for any $i \in [n]$, with $d$ the model width. Finally, a pooling operation reduces the sequence of output states to output encodings with a fixed size $\vy \in \R^M$. Examples of pooling operations include attentional pooling~\cite{lee2019set}, global average pooling, and extracting the \texttt{CLS} or \texttt{EOS} token representations for images or text respectively.

\subsection{Separate-head attention read-out (\Sparo)}
\label{sec:sparo}

We replace the pooling operation of transformer encoders with a concatenation of outputs of $L$ single-head attention mechanisms. \Sparo acts on the transformer outputs $\mH = \{\vh_1,\ldots,\vh_n\} \in \R^{n\times d}$ to produce the encoder output $\vy = \concat(\vy_1,\ldots,\vy_L) \in \R^{LV}$, where \Sparo slot $\vy_l \in \R^V$, $l \in [L]$. We illustrate the architecture of \Sparo in \cref{fig:sparo}. Each \Sparo slot is produced as:
\begin{align}
    \vy_l &= \mW \mK_l \mH^\T \softmax\left(\frac{\mH \mK_l^\T \vq_l}{\sqrt{D}}\right).\label{eq:sparo}
\end{align}
Here, $\mK_l \in \R^{D\times d}$ is a learned key projection weight, and $\vq_l \in \R^D$ is a learned query embedding for attention. We parameterize the value projection as $\mW \mK_l$ where $\mW \in \R^{V\times D}$ is learned parameter shared between all $L$ slots. Decomposing the value projection in this manner helps reduce the number of model parameters, allowing a larger choice of $L$ with fixed resources.

Generally, we pick $V$ and $D$ to be significantly smaller than the transformer width. In practice, we use the typical value of $D=64$ used in standard multi-head attention modules, and set $V$ to the same value in our experiments. Therefore, each \Sparo slot has limited expressivity and representational capacity. The bottleneck underpinning each \Sparo slot is a result of these constraints and the mechanistic bottleneck of a single attention mechanism instantiating competition between its input features.

\Sparo can be thought of as multi-head attention \cite{vaswani2017attention} where the queries are learned embeddings, key and value projection weights are shared, and the output projection weight is block-diagonal with each block containing the same parameters. Assuming $L=V=D=\sqrt{d}$, \Sparo requires a total of $d^2 + 2d$ parameters, compared to multi-head attention's $4d^2$.

\subsection{CLIP with \Sparo}
\label{sec:clipsparo}

We add \Sparo to both the image and text encoders of CLIP \cite{radford2021learning}, setting the same values of $L$ and $V$ in both encoders. Each \Sparo slot is $\ell_2$-normalized separately, and we divide their concatenation by $\sqrt{L}$ to produce the global $\ell_2$-normalized encoding. With this scheme, the cosine similarity between the image encoding $\vy^\text{i}$ and the text encoding $\vy^\text{t}$ becomes the expected cosine similarity between corresponding slot encodings:
\begin{align}
    {\vy^\text{i}}^\top\vy^\text{t}
    = \sum_{l=1}^L \norm{\vy^\text{i}_l}\norm{\vy^\text{t}_l} \frac{{\vy^\text{i}_l}^\top\vy^\text{t}_l}{\norm{\vy^\text{i}_l}\norm{\vy^\text{t}_l}}
    = \sum_{l=1}^L \frac{1}{L} \cosim(\vy^\text{i}_l, \vy^\text{t}_l)
    = \E_{l\sim [L]}\left[\cosim(\vy^\text{i}_l, \vy^\text{t}_l)\right].
\end{align}

We train the model using the standard CLIP loss which maximizes $\cosim(\vy^\text{i}, \vy^\text{t})$ of encodings $\vy^\text{i}$ and $\vy^\text{t}$ for aligned image-text pairs and minimizes it across all other pairings. With \Sparo's separate-head attention bottleneck, maximizing $\E_{l\sim [L]}\left[\cosim(\vy^\text{i}_l, \vy^\text{t}_l)\right]$ for aligned pairs across the training dataset encourages both modalities to learn similar attention semantics per slot $l \in [L]$. Consequently, \Sparo imparts a prior that both modalities encode a shared world with $L$ selectively attendable concepts, and similarities between inputs can be expressed as their average similarity for these concepts.

\subsection{DINO with \Sparo}
\label{sec:dinosparo}

The online and momentum encoders of DINO~\cite{caron2021emerging} are each comprised of a ViT backbone which produces the encoding, and a DINO head that transforms it into a categorical distribution for distillation. Unlike CLIP, these encoders have an inherent bias for learning similar encoding functions due to one being an exponentially moving average of the other. However, in standard DINO, there are no constraints for selective attention in the encoding structure. We add \Sparo to DINO to impart the prior for representing separately-attendable concepts in its encodings. Concretely, we replace the final transformer block of the ViT backbone with the \Sparo, but do not modify the DINO head.

\section{Results}
\label{sec:exp:benchmarks}

In \cref{sec:exp:benchmarks,sec:exp:analysis}, we evaluate our method on a variety of downstream tasks, analyze \Sparo's representation structure with the ability to intervene on selected slots, perform ablations to support our design, and visualize the learned concepts.

In this section, we validate that partitioning the representation structure as a collection of separately-attended concepts leads to improved generalization for downstream recognition, robustness, compositionality, and retrieval, using \Sparo with CLIP~\cite{radford2021learning} in \cref{sec:exp:acc,sec:exp:retrieval,sec:exp:linprobe} and DINO~\cite{caron2021emerging} in \cref{sec:exp:linprobe}.

\paragraph{Datasets.} We train our CLIP models on one of Conceptual Captions 3M (CC3M)~\cite{sharma2018conceptual}, Conceptual Captions 12M (CC12M)~\cite{changpinyo2021cc12m}, a combination of CC3M and CC12M (CC15M), or LAION-400M (L400M)~\cite{schuhmann2021laion}. \cref{sec:hyperparameters} provides the statistics for our training datasets.

\paragraph{Models.} For all experiments in this section, we train CLIP models using the open-source OpenCLIP~\cite{openclip} project. We consider two model sizes: those with a ViT-B/16 visual backbone (CLIP$^{16}$) and those with a ViT-B/32 backbone (CLIP$^{32}$). For models trained with \Sparo, we replace the last transformer block of the base transformer backbones. For our chosen settings, this ensures that the resulting model size is comparable to the original model size. Further, since only the \texttt{CLS} token output from the image transformer and the \texttt{EOS} token output from the text transformer are used for the standard CLIP encodings, removing one transformer block ensures that we do not gain an unfair advantage by attending to positions from a layer that standard CLIP discards. Note that standard CLIP also retains the MLP as a part of its final transformer block, which is absent in the \Sparo module. We also compare our encodings with those produced by CLIP with global average pooling (CLIP+GAP), which enjoys the advantage of being able to use all of the final layer outputs. When using \Sparo, we use $L=V=64$ when training on Conceptual Captions, and $L=128,V=64$ when training on LAION-400M. We also provide results for CLIP with residual network (ResNet) \cite{he2016deep} encoders in \cref{sec:resnets}.

\subsection{Zero-shot recognition, robustness, and compositionality}
\label{sec:exp:acc}

\begin{table}[tb]
    \caption{Zero-shot recognition, robustness, and SugarCrepe compositionality for CLIP models trained on Conceptual Captions and LAION-400M.}
    \label{tab:robustness}
    \centering
    \begin{tabular}{@{}llccccccc@{}}
        \toprule
        \multirow{2}[2]{*}{Train} &\multirow{2}[2]{*}{Model} &\multicolumn{5}{c}{ImageNet-} &\multirow{2}[2]{*}{\makecell{Object\\Net}} &\multirow{2}[2]{*}{\makecell{Sugar\\Crepe}} \\\cmidrule{3-7}
        & &V1 &V2 &Sketch &A &R & & \\
        \midrule
        \multirow{3}{*}{CC3M} &CLIP$^{16}$ $^{(\mathcal{C})}$ &0.141 &0.122 &0.068 &0.033 &0.177 &0.080 &0.611 \\
        &$\mathcal{C}$+GAP &0.156 &0.134 &0.069 &0.033 &0.187 &0.081 &0.616 \\
        &$\mathcal{C}$+\Sparo &\textbf{0.170} &\textbf{0.140} &\textbf{0.088} &\textbf{0.035} &\textbf{0.221} &\textbf{0.098} &\textbf{0.625} \\
        \midrule
        \multirow{3}{*}{CC12M} &CLIP$^{16}$ $^{(\mathcal{C})}$ &0.361 &0.311 &0.249 &0.091 &0.467 &0.218 &0.697 \\
        &$\mathcal{C}$+GAP &0.382 &0.330 &0.262 &0.101 &0.501 &0.241 &0.695 \\
        &$\mathcal{C}$+\Sparo &\textbf{0.406} &\textbf{0.350} &\textbf{0.298} &\textbf{0.113} &\textbf{0.559} &\textbf{0.268} &\textbf{0.723} \\
        \midrule
        \multirow{3}{*}{CC15M} &CLIP$^{16}$ $^{(\mathcal{C})}$ &0.384 &0.337 &0.268 &0.105 &0.503 &0.238 &0.699 \\
        &$\mathcal{C}$+GAP &0.399 &0.343 &0.287 &0.114 &0.531 &0.252 &0.701 \\
        &$\mathcal{C}$+\Sparo &\textbf{0.437} &\textbf{0.378} &\textbf{0.317} &\textbf{0.145} &\textbf{0.579} &\textbf{0.279} &\textbf{0.730} \\
        \midrule
        \multirow{3}{*}{L400M} &CLIP$^{32}$ $^{(\mathcal{C})}$ &0.617 &0.531 &0.482 &0.202 &0.719 &0.423 &0.748 \\
        &$\mathcal{C}$+GAP &0.623 &0.537 &0.492 &0.212 &0.725 &0.440 &0.732 \\
        &$\mathcal{C}$+\Sparo &\textbf{0.635} &\textbf{0.552} &\textbf{0.507} &\textbf{0.231} &\textbf{0.747} &\textbf{0.459} &\textbf{0.770} \\
        \bottomrule
    \end{tabular}
\end{table}

We evaluate the zero-shot classification accuracy of trained CLIP models on ImageNet \cite{deng2009imagenet} and a set of robustness benchmarks including ImageNet-V2 \cite{recht2019imagenet}, ImageNet-Sketch \cite{wang2019learning}, ImageNet-A \cite{hendrycks2021natural}, ImageNet-R \cite{hendrycks2021many}, and ObjectNet \cite{barbu2019objectnet}. Additionally, to evaluate the compositionality of our learned encodings in terms of objects, attributes, and relations, we test our trained models on SugarCrepe \cite{hsieh2023sugarcrepe}. We report our results in \cref{tab:robustness}, and provide more fine-grained SugarCrepe numbers in \cref{sec:finesugarcrepe}. We find that using \Sparo encodings outperforms CLIP and CLIP+GAP in all settings considered. The improved results on the SugarCrepe benchmark suggests that \Sparo encodings exhibit compositionality at a level that enable better binding of object, attribute, and relation properties than baseline encodings on average.

\begin{figure}[tb]
    \centering
    \hfill\begin{subfigure}{0.475\linewidth}
        \centering
        \includegraphics[width=\linewidth]{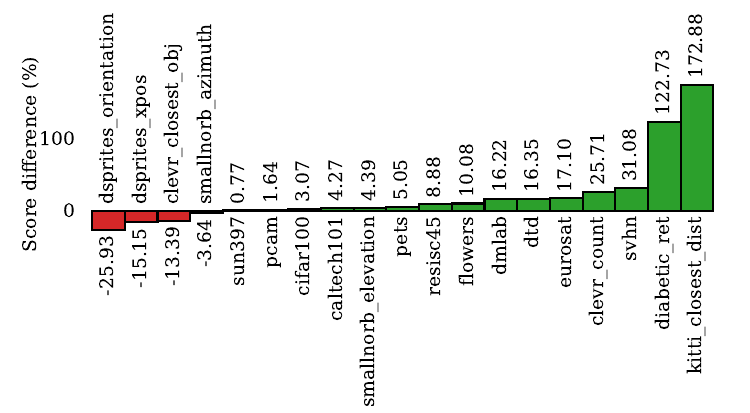}
        \caption{CLIP$^{16}$ models trained on CC3M.}
        \label{fig:vtabcc3m}
    \end{subfigure}
    \hfill\begin{subfigure}{0.475\linewidth}
        \centering
        \includegraphics[width=\linewidth]{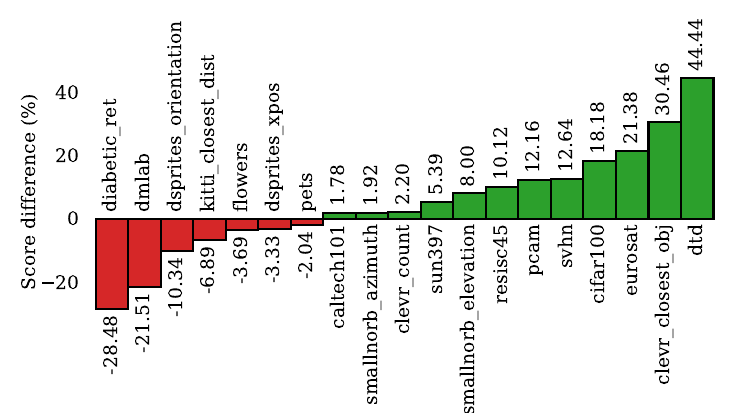}
        \caption{CLIP$^{16}$ models trained on CC12M.}
        \label{fig:vtabcc12m}
    \end{subfigure}
    \hfill\hfill\\
    \hfill\begin{subfigure}{0.475\linewidth}
        \centering
        \includegraphics[width=\linewidth]{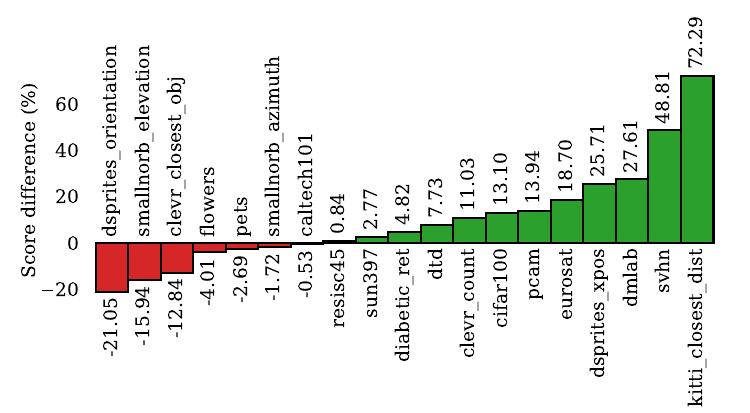}
        \caption{CLIP$^{16}$ models trained on CC15M.}
        \label{fig:vtabcc15m}
    \end{subfigure}
    \hfill\begin{subfigure}{0.475\linewidth}
        \centering
        \includegraphics[width=\linewidth]{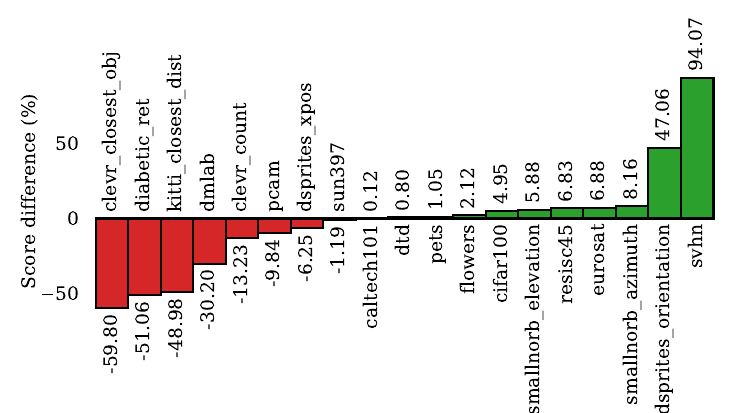}
        \caption{CLIP$^{32}$ models trained on L400M.}
        \label{fig:vtablaion400m}
    \end{subfigure}
    \hfill\hfill%
    \caption{Relative differences of CLIP+\Sparo zero-shot accuracies when compared to CLIP+GAP on the VTAB benchmark.}
    \label{fig:vtab}
\end{figure}

Finally, we also use the VTAB benchmark \cite{zhai2019large} to evaluate \Sparo on a diverse set of datasets \cite{fei2006one,krizhevsky2009learning,cimpoi2014describing,nilsback2008automated,parkhi2012cats,xiao2010sun,yuval2011reading,helber2019eurosat,cheng2017remote,veeling2018rotation,dugas2015diabetic,johnson2017clevr,matthey2017dsprites,lecun2004learning,beattie2016deepmind,geiger2013vision} that vary more significantly from the training data. We present the relative zero-shot classification accuracy improvements of CLIP+\Sparo over CLIP+GAP in \cref{fig:vtab} and the corresponding absolute values in \cref{sec:absvtab}. We see that \Sparo outperforms standard CLIP encodings for a majority of VTAB tasks in all four settings.

\subsection{Zero-shot image and text retrieval}
\label{sec:exp:retrieval}

\begin{table}[tb]
    \caption{Image and text zero-shot Recall@5 retrieval results on MS~COCO, Flickr8k, and Flickr30k for CLIP models trained on Conceptual Captions and LAION-400M.}
    \label{tab:retrieval}
    \centering
    \begin{tabular}{@{}llcccp{0.17em}ccc@{}}
        \toprule
        \multirow{2}[2]{*}{Train} &\multirow{2}[2]{*}{Model} &\multicolumn{3}{c}{Image R@5} & &\multicolumn{3}{c}{Text R@5} \\\cmidrule{3-5}\cmidrule{7-9}
        & &COCO &F8k &F30k & &COCO &F8k &F30k \\\midrule
        \multirow{3}{*}{CC3M} &CLIP$^{16}$ $^{(\mathcal{C})}$ &0.237 &0.400 &0.353 & &0.294 &0.489 &0.469 \\
        &$\mathcal{C}$+GAP &0.260 &0.430 &0.400 & &0.312 &0.534 &0.497 \\
        &$\mathcal{C}$+\Sparo &\textbf{0.289} &\textbf{0.490} &\textbf{0.461} & &\textbf{0.363} &\textbf{0.587} &\textbf{0.581} \\
        \midrule
        \multirow{3}{*}{CC12M} &CLIP$^{16}$ $^{(\mathcal{C})}$ &0.474 &0.691 &0.715 & &0.614 &0.837 &0.817 \\
        &$\mathcal{C}$+GAP &0.499 &0.714 &0.725 & &0.628 &0.819 &0.847 \\
        &$\mathcal{C}$+\Sparo &\textbf{0.526} &\textbf{0.737} &\textbf{0.759} & &\textbf{0.670} &\textbf{0.878} &\textbf{0.878} \\
        \midrule
        \multirow{3}{*}{CC15M} &CLIP$^{16}$ $^{(\mathcal{C})}$ &0.512 &0.738 &0.757 & &0.636 &0.844 &0.865 \\
        &$\mathcal{C}$+GAP &0.535 &0.761 &0.772 & &0.659 &0.849 &0.881 \\
        &$\mathcal{C}$+\Sparo &\textbf{0.557} &\textbf{0.778} &\textbf{0.793} & &\textbf{0.696} &\textbf{0.898} &\textbf{0.905} \\
        \midrule
        \multirow{3}{*}{L400M} &CLIP$^{32}$ $^{(\mathcal{C})}$ &0.599 &0.822 &0.840 & &0.765 &0.923 &0.935 \\
        &$\mathcal{C}$+GAP &0.610 &0.822 &0.845 & &0.768 &0.917 &0.946 \\
        &$\mathcal{C}$+\Sparo &\textbf{0.616} &\textbf{0.836} &\textbf{0.854} & &\textbf{0.774} &\textbf{0.935} &\textbf{0.950} \\
        \bottomrule
    \end{tabular}
\end{table}

To separately evaluate the quality of the learned image and text encodings, we consider zero-shot retrieval based on image and text on MS~COCO \cite{lin2014microsoft}, Flickr8k \cite{hodosh2013framing}, and Flickr30k \cite{young2014image}. Our results are presented in \cref{tab:retrieval}. To compute Recall@5 for `Text' we find the nearest 5 texts for an image, calculate the fraction of those texts that match the image, and average this metric over all the images (similar for Recall@5 for `Image'). We find that \Sparo encodings improve retrieval across all three datasets for both modalities.

\subsection{Linear probe and DINO nearest neighbors classification}
\label{sec:exp:linprobe}

\begin{table}[tb]
    \caption{\textbf{Left:} ImageNet linear probe accuracy for CLIP models trained on Conceptual Captions and LAION-400M. \textbf{Right:} ImageNet $20$-nearest neighbors and linear probe accuracy for encodings trained with DINO.}
    \label{tab:linprobe}
    \centering
    \hfill\begin{tabular}{@{}lccccc@{}}\toprule
        \multirow{2}[2]{*}{Model} &\multicolumn{4}{c}{ImageNet linear probe} \\\cmidrule{2-6}
        &Train: &CC3M &CC12M &CC15M &L400M \\\midrule
        \multicolumn{2}{l}{CLIP $^{(\mathcal{C})}$} &0.469 &0.630 &0.646 &0.743 \\
        \multicolumn{2}{l}{$\mathcal{C}$+GAP} &0.504 &0.649 &0.664 &0.747 \\
        \multicolumn{2}{l}{$\mathcal{C}$+\Sparo} &\textbf{0.561} &\textbf{0.700} &\textbf{0.711} &\textbf{0.755} \\
        \bottomrule
    \end{tabular}
    \hfill\begin{tabular}{@{}lcc@{}}
        \toprule
        \multirow{2}[2]{*}{Model} &\multicolumn{2}{c}{ImageNet eval} \\\cmidrule{2-3}
        &$20$-NN &Linear probe \\
        \midrule
        DINO &0.685 &0.735 \\
        DINO+\Sparo &\textbf{0.706} &\textbf{0.757} \\
        \bottomrule
    \end{tabular}
    \hfill\hfill%
\end{table}

We evaluate the classification accuracy of linear probes trained on the encodings of our trained models.
For CLIP models, we follow \cite{radford2021learning} where we take a subset of the training data to use as a validation set, and sweep for the best weight decay hyperparameter on it for each model setting. Additionally, we evaluate \Sparo for self-supervised learning in vision using DINO \cite{caron2021emerging}. We train DINO with a ViT-S/16 backbone on ImageNet \cite{deng2009imagenet} without labels with 8 GPUs for 100 epochs without changing any other hyperparameters. Here, we use \Sparo with $L=V=64$, replacing the last transformer block of the ViT encoder. The trained DINO models are evaluated on $k$-nearest neighbours with $k=20$ and linear classification on ImageNet with labels. We report our results for CLIP and DINO in \cref{tab:linprobe}. Our results for CLIP align with our previous results, with \Sparo outperforming CLIP and CLIP+GAP in all linear probe settings. Furthermore, we see that a straightforward incorporation of \Sparo in the DINO encoder outperforms standard DINO for both nearest neighbors and linear classification. We provide additional experimental details in \cref{sec:cliphyperparams,sec:dinohyperparams}.

\section{Analysis}
\label{sec:exp:analysis}

In this section, we provide insights into encodings learned by \Sparo. We demonstrate the ability of using \Sparo's slot structure to perform post-hoc selection of slots for improving downstream performance. We then show that without this structure, such selection is more prone to overfitting. Next, we validate the design choice of using single-head attention to model the notion of selective attention through controlled ablations. Finally, we visualize example concepts that CLIP models trained with \Sparo learn to attend to.

\subsection{Post-hoc concept selection}
\label{sec:exp:naiveslots}

\begin{table}[tb]
    \caption{Effect of intervening to selecting only the top $32$ slots based on zero-shot ImageNet validation accuracy on SugarCrepe compositionality for CC15M-trained CLIP$^{16}$+\Sparo and LAION-400M-trained CLIP$^{32}$+\Sparo.}
    \label{tab:selectslots}
    \centering
    \begin{tabular}{@{}lccccp{0.17em}cc@{}}\toprule
        \multicolumn{2}{c}{\multirow{2}[2]{*}{SugarCrepe}} &\multicolumn{3}{c}{CC15M} & &\multicolumn{2}{c}{L400M} \\\cmidrule{3-5}\cmidrule{7-8}
& & Slots: &$100\%$ &$50\%$ & &$100\%$ &$25\%$ \\
        \midrule
        \multirow{3}{*}{Replace} &\multicolumn{2}{l}{Object} &0.896 &\textbf{0.912} & &0.927 &\textbf{0.929} \\
        &\multicolumn{2}{l}{Attribute} &\textbf{0.798} &0.797 & &0.822 &\textbf{0.854} \\
        &\multicolumn{2}{l}{Relation} &0.666 &\textbf{0.713} & &0.678 &\textbf{0.705} \\
        \cmidrule{1-3}
        \multirow{2}{*}{Swap} &\multicolumn{2}{l}{Object} &0.593 &\textbf{0.659} & &0.585 &\textbf{0.646} \\
        &\multicolumn{2}{l}{Attribute} &0.700 &\textbf{0.703} & &0.719 &\textbf{0.740} \\
        \cmidrule{1-3}
        \multirow{2}{*}{Add} &\multicolumn{2}{l}{Object} &0.779 &\textbf{0.809} & &\textbf{0.866} &0.857 \\
        &\multicolumn{2}{l}{Attribute} &0.679 &\textbf{0.737} & &0.794 &\textbf{0.822} \\
        \cmidrule{1-3}
        \multicolumn{3}{c}{Average} &0.730 &\textbf{0.761} & &0.770 &\textbf{0.793} \\
        \bottomrule
    \end{tabular}
\end{table}

\Sparo enables downstream tasks access to a collection of concepts. However, not all concepts may be important for a specific downstream task. For example, attending to watermarks or caption style can be important when trying to determine the source of a sample, but is unnecessary and often harmful for most downstream tasks that focus on the elements in the scenes. The separation of concepts in the structure of a \Sparo encoding makes it possible to manually intervene to select desirable slots.

We demonstrate this ability with a simple heuristic: we perform concept selection by computing the zero-shot ImageNet validation accuracy of each slot separately, and pick only the top-performing $32$ slots. In \cref{tab:selectslots}, we show that such an intervention exposes improved compositionality when evaluating CLIP+\Sparo models trained on CC15M and LAION-400M on SugarCrepe.

\subsection{Robustness of \Sparo's slot structure}
\label{sec:exp:masking}

\begin{table}[tb]
    \caption{Effect of overfitting a partitioned mask for image encodings $\vy$ of frozen CC15M-trained CLIP$^{32}$+GAP and CLIP$^{32}$+\Sparo models by training on SugarCrepe. We normalize the relative change on the evaluation benchmarks by the attained relative change on SugarCrepe. The \hl{\,highlighted\,} rows indicate settings where the structure assumed for the mask aligns with \Sparo's representation structure.}
    \label{tab:masking}
    \centering
    \begin{tabular}{@{}lllccp{0.17em}cccccc@{}}
        \toprule
        \multirow{3}[7]{*}{\makecell{(\# parts,\\Part size)}} &\multirow{3}[7]{*}{\makecell{Model\\(CLIP$^{32}$)}} &\multirow{3}[7]{*}{\makecell{Shape\\of $\vy$}} &\multicolumn{2}{c}{Training} & &\multicolumn{6}{c}{Evaluation} \\\cmidrule{4-5}\cmidrule{7-12}
        & & &\multicolumn{2}{c}{SugarCrepe} & &\multicolumn{3}{c}{VTAB Avg} &\multicolumn{3}{c}{ImageNet $0$-shot Acc} \\
        \cmidrule{4-5}\cmidrule{7-12}
        & & &{Initial} &{Masked} & &{Initial} &{Masked} &{\,\makecell{Relative\\change}\,} &{Initial} &{Masked} &{\,\makecell{Relative\\change}} \\
        \midrule
        \multirow{2}{*}{(64, 64)} &+GAP &(4096) &\scriptsize 0.692 &\scriptsize 0.732 & &\scriptsize 0.240 &\scriptsize 0.217 &$-$1.689 &\scriptsize 0.343 &\scriptsize 0.292 &$-$2.598 \\
        &\cellh+\Sparo &\cellh(64, 64) &\scriptsize \cellh0.710 &\scriptsize \cellh0.764 &\cellh &\scriptsize \cellh0.262 &\scriptsize \cellh0.272 &\cellh\textbf{$+$0.497} &\scriptsize \cellh0.370 &\scriptsize \cellh0.364 &\cellh\textbf{$-$0.190} \\
        \midrule
        \multirow{2}{*}{(64, 8)} &+GAP &(512) &\scriptsize 0.678 &\scriptsize 0.712 & &\scriptsize 0.254 &\scriptsize 0.219 &$-$2.794 &\scriptsize 0.345 &\scriptsize 0.279 &$-$3.924 \\
        &\cellh+\Sparo &\cellh(64, 8) &\scriptsize \cellh0.696 &\scriptsize \cellh0.741 &\cellh &\scriptsize \cellh0.249 &\scriptsize \cellh0.260 &\cellh\textbf{$+$0.688} & \scriptsize\cellh0.355 &\scriptsize \cellh0.347 &\cellh\textbf{$-$0.347} \\
        \midrule
        \multirow{2}{*}{(512, 1)} &+GAP &(512) &\scriptsize 0.678 &\scriptsize 0.771 & &\scriptsize 0.254 &\scriptsize 0.215 &$-$1.119 &\scriptsize 0.345 &\scriptsize 0.250 &\textbf{$-$2.003} \\
        &+\Sparo &(64, 8) &\scriptsize 0.696 &\scriptsize 0.809 & &\scriptsize 0.249 &\scriptsize 0.206 &\textbf{$-$1.059} &\scriptsize 0.355 &\scriptsize 0.224 &$-$2.277 \\
        \midrule
        \multirow{2}{*}{(4096, 1)} &+GAP &(4096) &\scriptsize 0.692 &\scriptsize 0.815 & &\scriptsize 0.240 &\scriptsize 0.215 &$-$0.596 &\scriptsize 0.343 &\scriptsize 0.286 &$-$0.938 \\
        &+\Sparo &(64, 64) &\scriptsize 0.710 &\scriptsize 0.881 & &\scriptsize 0.262 &\scriptsize 0.242 &\textbf{$-$0.316} &\scriptsize 0.370 &\scriptsize 0.328 &\textbf{$-$0.465} \\
        \bottomrule
    \end{tabular}
\end{table}

We further utilize the notion of post-hoc concept selection to study the robustness of \Sparo's representational structure, and compare it with that of standard encodings. We set up the experiment as an inverse of \cref{sec:exp:naiveslots} --- we perform concept selection using the smaller SugarCrepe benchmark and evaluate the resulting encodings on real-world benchmarks. SugarCrepe contains 7,512 example pairs of positive and negative captions, differing only in compositional interpretation in 7 ways, over a set of 1,561 images. Compositionality is a shared high-level notion present in most real-world data, yet there is much information that exists in data beyond what is necessary to solve the SugarCrepe tasks. The small size of the SugarCrepe as a training dataset, together with the ubiquitous nature of compositionality in machine learning tasks, makes it an ideal candidate to overfit concept selection to for stress-testing the representational structure of \Sparo by evaluating on other tasks.

We train a global mask $\vm \in [0,1]^M$ of the image encodings $\vy \in \R^M$ to maximize SugarCrepe performance, and evaluate the masked encodings $\vm \odot \vy$ on VTAB tasks and ImageNet for zero-shot classification. We train masks for encodings of frozen CLIP$^{32}$+GAP and CLIP$^{32}$+\Sparo models trained on CC15M with $M=512$ and $M=4096$ each. For the \Sparo models, we use $L=64$ and $V=M/L$. For each of these settings, we consider two types of masking: per-dimension and per-slot. To compare the CLIP$^{32}$+GAP encodings on per-slot masking, we consider their $L$ contiguous equal-sized partitions as the slot encodings. We provide full implementation details of our mask training setup in \cref{sec:maskingdetails}.

We expect the compositionality of the data to be encoded in concepts that are useful for real-world downstream benchmarks. Therefore, the aggressive selection of only the concepts that perform well on a compositionality benchmark should not significantly impact downstream performance. However, since SugarCrepe does not represent all possible compositional manipulations, we expect to see a drop in performance on some tasks due to overfitting. We present our results in \cref{tab:masking}. We find that when the mask structure is aligned with the slot structure of \Sparo, overfitting concept selection improves the average performance on VTAB tasks, and incurs only a minimal drop in performance on ImageNet. Furthermore, even in settings where the masks are not trained to align with \Sparo's slot structure, we see that \Sparo remains more resilient to overfitting than standard encodings in a majority of settings.

\subsection{Ablating the bottleneck and cross-modal alignment of \Sparo}
\label{sec:exp:singlehead}

\begin{table}[tb]
    \caption{Ablation experiments showing benefits of \Sparo's attention bottleneck (Cross-attention is `Separate-head') and cross-modal attention alignment (Slot-wise LayerNorm and projection are `Yes' or `N/A') with CC15M-trained CLIP, CLIP+GAP, CLIP+\Sparo, and ablated variants of CLIP+\Sparo that include multi-head cross-attention and the attentional pooler (AttPool) \cite{lee2019set,yu2022coca}.}
    \label{tab:attnpool}
    \centering
    \begin{tabular}{@{}lccccccc@{}}
        \toprule
        Model &\makecell{Replace\\last block} &\makecell{Cross-\\attention} &\makecell{Slot-wise\\LayerNorm} &\makecell{Slot-wise\\projection} &Params &\makecell{\textsc{flops}\\($\times 10^9$)} &\makecell{ImageNet\\zero-shot} \\
        \midrule
        CLIP$^{32}$ $^{(\mathcal{C})}$ &\multirow{2}{*}{N/A} &\multirow{2}{*}{N/A} &\multirow{2}{*}{N/A} &\multirow{2}{*}{N/A} &151M &7.4 &0.329 \\
        $\mathcal{C}$+GAP & & & & &151M &7.4 &0.345 \\
        \midrule
        $\mathcal{C}$+AttPool &\multirow{9}[6]{*}{No} &\multirow{4}[3]{*}{Multi-head} &\multirow{2}{*}{No} &No &440M &8.9 &0.312 \\\cmidrule{1-1}
        \multirow{7}[4]{*}{\makecell[l]{Ablated\\$\mathcal{C}$+\Sparo}}& & & &Yes &306M &8.8 &0.344 \\
        \cmidrule{4-5}
        & & &\multirow{2}{*}{Yes} &No &440M &8.9 &0.338 \\
        & & & &Yes &306M &8.8 &0.359 \\
        \cmidrule{3-5}
        & &\multirow{6}[4]{*}{Separate-head} &\multirow{2}{*}{No} &No &306M &8.8 &0.344 \\
        & & & &Yes &172M &8.7 &0.358 \\
        \cmidrule{4-5}
        & & &\multirow{2}{*}{Yes} &No &306M &8.8 &0.344 \\
        & & & &Yes &172M &8.7 &\textbf{0.374} \\\cmidrule{1-1}\cmidrule{4-5}
        \multirow{2}[2]{*}{$\mathcal{C}$+\Sparo} & & &\multirow{2}[2]{*}{N/A} &\multirow{2}[2]{*}{N/A} &161M &8.0 &0.372 \\\cmidrule{2-2}
        &Yes & & & &151M &7.4 &0.370 \\
        \bottomrule
    \end{tabular}
\end{table}

We evaluate the generalization benefits of the separate-head attention bottleneck and of aligning the \Sparo attention heads between the modalities in CLIP. We set up \Sparo on one end of the ablation spectrum and the cross-attention read-out of attentional pooler (AttPool) \cite{lee2019set,yu2022coca} to the other. AttPool uses multi-head attention to attend to the backbone encoder's outputs using embedded learned queries whereas \Sparo uses separate single-head attention mechanisms. Additionally, AttnPool performs layer normalization \cite{ba2016layer} followed by a linear projection on the output of attention to produce the output encoding, whereas \Sparo directly uses the output of separate-head attention as the encoding.

We train CLIP$^{32}$, CLIP$^{32}$+GAP, and variants of CLIP$^{32}$+\Sparo on CC15M with $L=128,V=64$ and evaluate for ImageNet zero-shot accuracy and model size. We start by considering the choice of either separate-head or multi-head cross-attention over the backbone outputs. For each cross-attention setting, we apply layer normalization and linear projection operations to the attention module's outputs, similar to AttnPool. However, we additionally also consider slot-wise variants of these operations. Here, a slot-wise operation $f_\phi$ is one that acts separately on each input slot $\vy_l$ using the same parameters $\phi$, and produces the slot-structured output $\concat(f_\phi(\vy_1), \ldots, f_\phi(\vy_L))$. When applied to an input that is not slot-structured, we assume a slot structure by partitioning the input into $L$ contiguous equal-sized partitions. With \Sparo, using both slot-wise layer normalization and slot-wise projection enables each read-out attention head from one modality to align with the corresponding head from the other without being influenced by the other heads.

We present our results in \cref{tab:attnpool}, showing a clear advantage of \Sparo over AttnPool. Additionally, both the choices of separate-head attention over multi-head and slot-wise operations over the alternatives result in improved generalization, reduced model size, and fewer floating-point operations per second (\textsc{flops}). While the former supports the effectiveness of the separate-head attention bottleneck, the latter validates the benefits of the prior for shared separately-attendable concepts between the modalities enabled by \Sparo.

\subsection{Visualizations}
\label{sec:exp:visualizations}

\begin{figure}[tb]
    \visspacing
    \centering
    \begin{subfigure}{0.26\linewidth}
        \centering
        \includegraphics[width=\linewidth,height=\visheight,keepaspectratio]{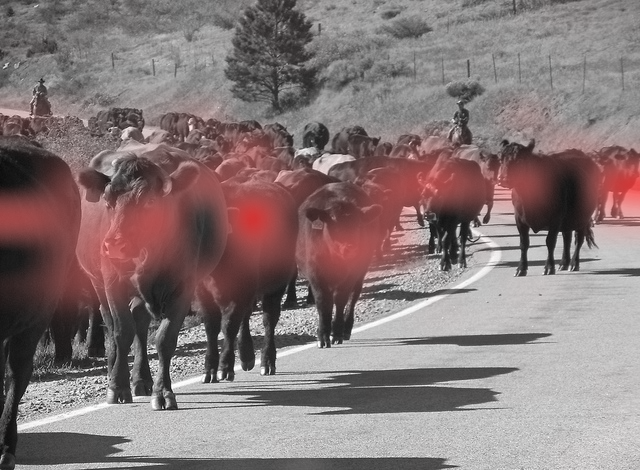}
        \caption*{\sffamily\viscaptionsize A \textcolor{BrickRed}{herd} of \textcolor{red}{\textbf{cattle}} walking down a road being followed by a cowboy}
    \end{subfigure}
    \begin{subfigure}{0.24\linewidth}
        \centering
        \includegraphics[width=\linewidth,height=\visheight,keepaspectratio]{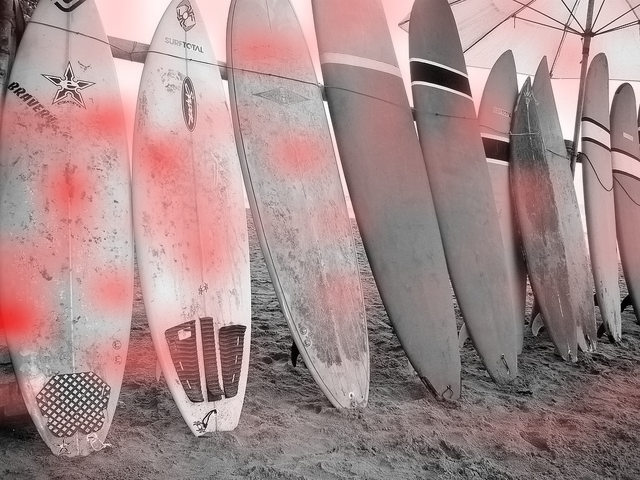}
        \caption*{\sffamily\viscaptionsize Several \textcolor{red}{surf\textbf{boards}} standing in a row on the beach}
    \end{subfigure}
    \begin{subfigure}{0.25\linewidth}
        \centering
        \includegraphics[width=\linewidth,height=\visheight,keepaspectratio]{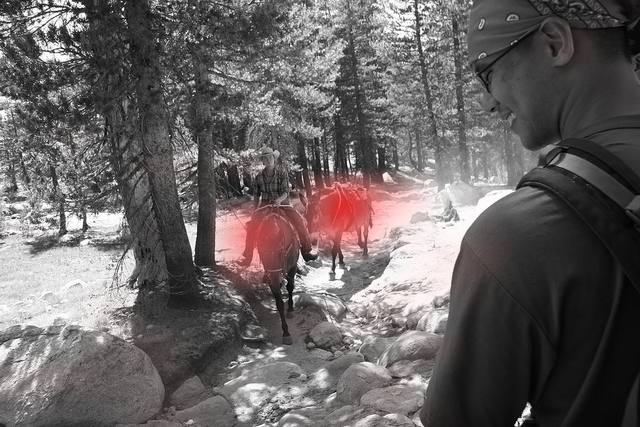}
        \caption*{\sffamily\viscaptionsize Two people riding \textcolor{red}{\textbf{horses}} on a \textcolor{BrickRed}{rock} path}
    \end{subfigure}
    \begin{subfigure}{0.22\linewidth}
        \centering
        \includegraphics[width=\linewidth,height=\visheight,keepaspectratio]{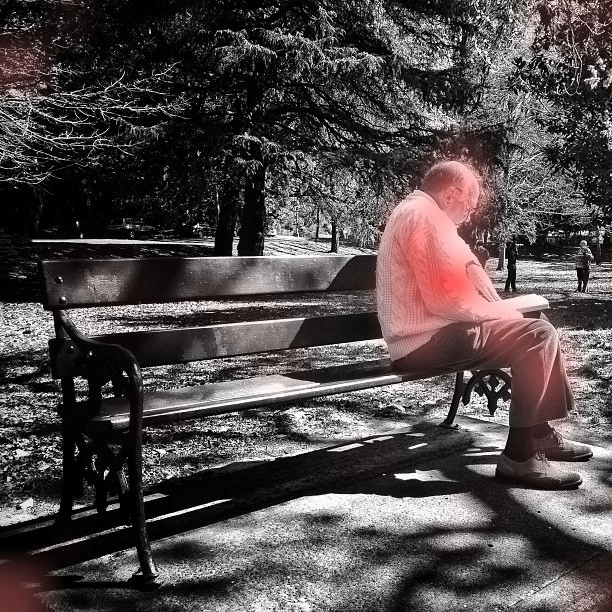}
        \caption*{\sffamily\viscaptionsize A \textcolor{red}{\textbf{man}} sitting alone on a park bench in a park}
    \end{subfigure}\\
    \begin{subfigure}{0.26\linewidth}
        \centering
        \includegraphics[width=\linewidth,height=\visheight,keepaspectratio]{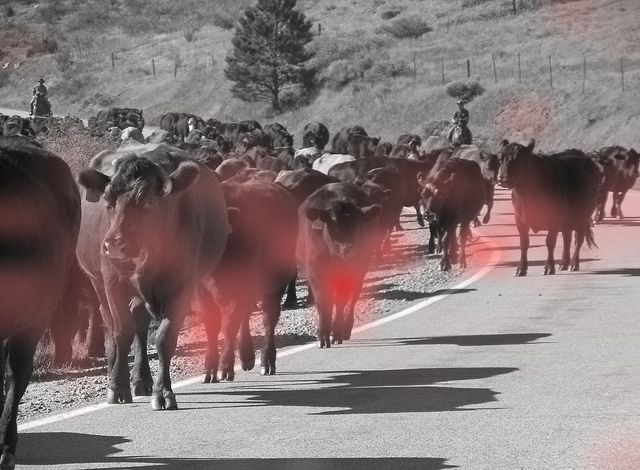}
        \caption*{\sffamily\viscaptionsize A herd of cattle \textcolor{red}{\textbf{walking} down} a road being followed by a cowboy}
    \end{subfigure}
    \begin{subfigure}{0.24\linewidth}
        \centering
        \includegraphics[width=\linewidth,height=\visheight,keepaspectratio]{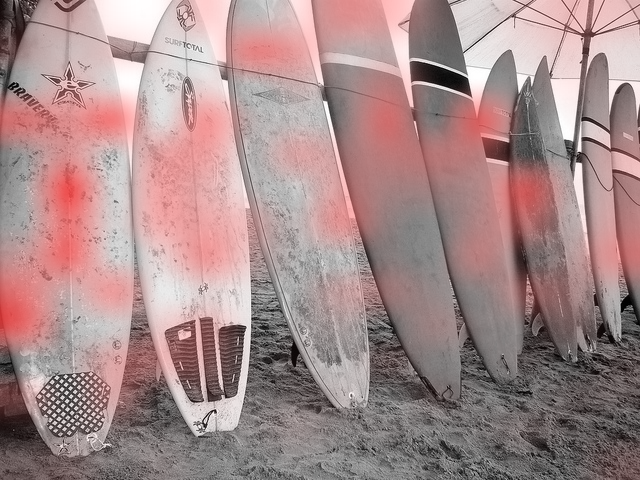}
        \caption*{\sffamily\viscaptionsize Several surf\textcolor{BrickRed}{boards} \textcolor{red}{\textbf{standing}} \textcolor{BrickRed}{in} a \textcolor{red}{row} \textcolor{BrickRed}{on} the beach}
    \end{subfigure}
    \begin{subfigure}{0.25\linewidth}
        \centering
        \includegraphics[width=\linewidth,height=\visheight,keepaspectratio]{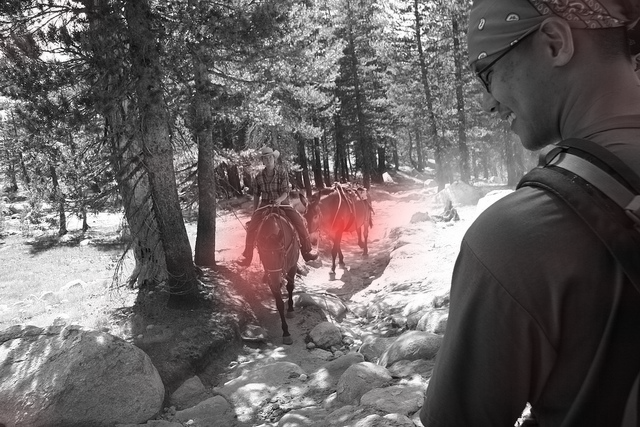}
        \caption*{\sffamily\viscaptionsize Two people \textcolor{red}{\textbf{riding} horses} on a rock path}
    \end{subfigure}
    \begin{subfigure}{0.22\linewidth}
        \centering
        \includegraphics[width=\linewidth,height=\visheight,keepaspectratio]{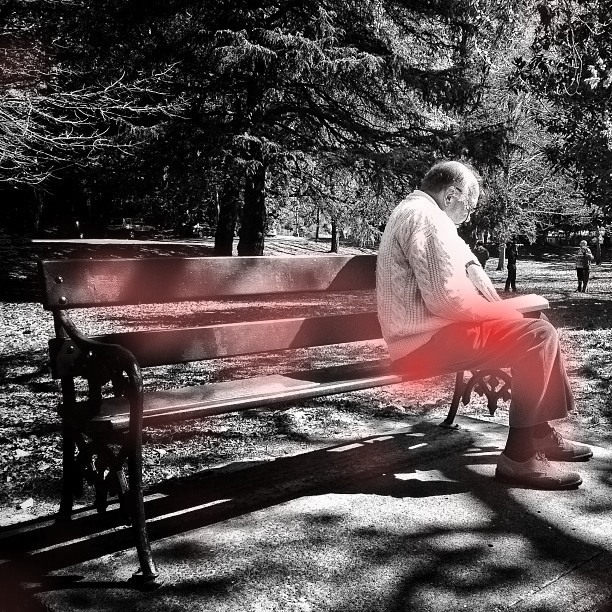}
        \caption*{\sffamily\viscaptionsize A man \textcolor{red}{sitting} alone on a park \textcolor{red}{\textbf{bench}} in a park}
    \end{subfigure}\\
    \begin{subfigure}{0.26\linewidth}
        \centering
        \includegraphics[width=\linewidth,height=\visheight,keepaspectratio]{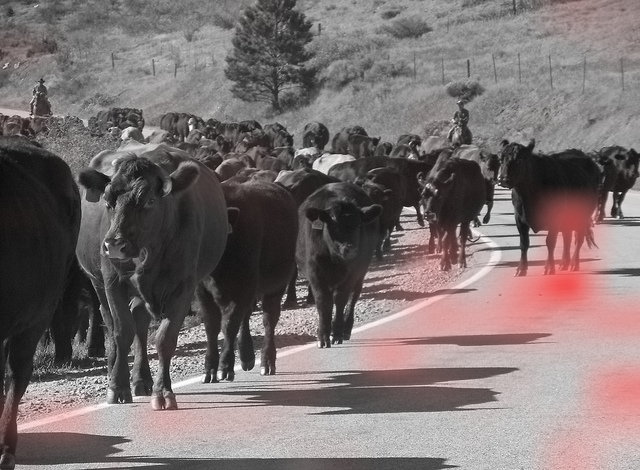}
        \caption*{\sffamily\viscaptionsize A herd of cattle walking down a \textcolor{red}{\textbf{road}} being followed by a cowboy}
    \end{subfigure}
    \begin{subfigure}{0.24\linewidth}
        \centering
        \includegraphics[width=\linewidth,height=\visheight,keepaspectratio]{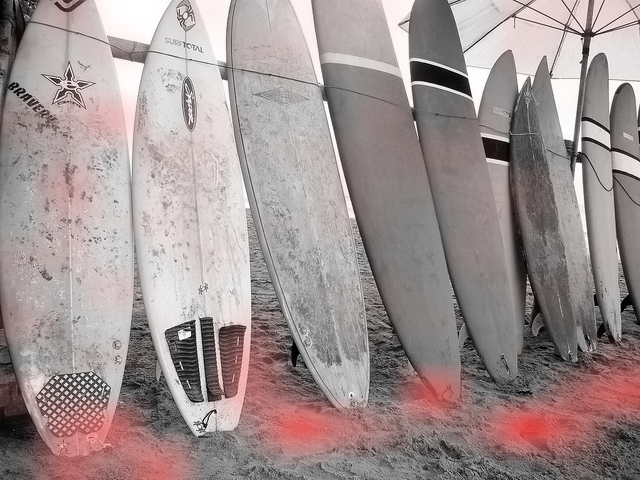}
        \caption*{\sffamily\viscaptionsize Several surfboards standing in a row on the \textcolor{red}{\textbf{beach}}}
    \end{subfigure}
    \begin{subfigure}{0.25\linewidth}
        \centering
        \includegraphics[width=\linewidth,height=\visheight,keepaspectratio]{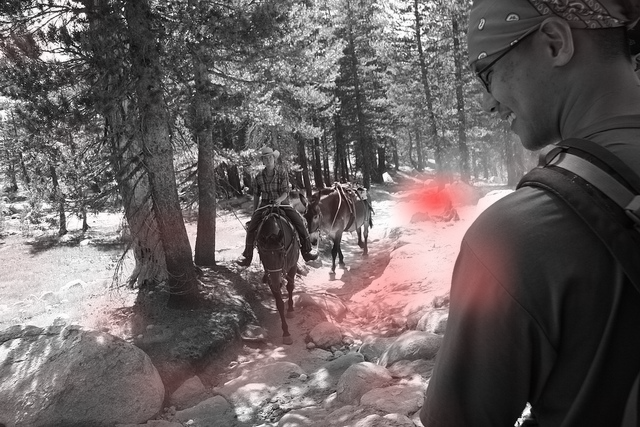}
        \caption*{\sffamily\viscaptionsize Two people riding horses on a \textcolor{red}{rock \textbf{path}}}
    \end{subfigure}
    \begin{subfigure}{0.22\linewidth}
        \centering
        \includegraphics[width=\linewidth,height=\visheight,keepaspectratio]{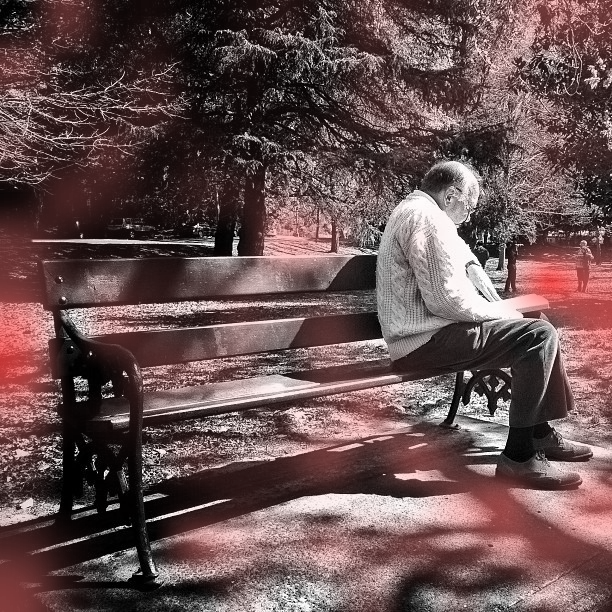}
        \caption*{\sffamily\viscaptionsize A man sitting alone on a \textcolor{BrickRed}{park} bench in a \textcolor{red}{\textbf{park}}}
    \end{subfigure}
    \caption{Visualizing of the \textcolor{red}{attended} image and text positions for three \Sparo slots (one per row) across four examples (one per column) from MS~COCO. We surmise that the \Sparo concepts from top to bottom represent the subject, activity, and location.}
    \label{fig:visualization}
\end{figure}

To conclude our analysis, we qualitatively visualize the concepts represented by \Sparo using the CLIP$^{16}$+\Sparo model trained on CC15M from \cref{sec:exp:acc}. A limitation of visualizing attended positions is that it only provides partial insight into the represented concept --- we can tell where the information is taken from, but not how the information is used. Different \Sparo slots can have similar attention maps for the same sample, but reveal different semantics when comparing patterns across other images. For instance, a slot that attends to animals (\eg dogs, cats, elephants) and another that attends to transportation (\eg trains, boats, cars) can both have the same attention mask over patches and words for a horse being used for transportation. Furthermore, not all \Sparo concepts align between the two modalities for every sample due to information being present in one that is not possible to infer from the other.

To ease our interpretation of attended concepts, we choose concepts to visualize that have sharp attention in the text modality, attend to non-overlapping text tokens, and have cosine similarity  more than 0.75 with the corresponding image slots. We present examples of these filtered concepts in \cref{fig:visualization}. In this example, we see that \Sparo is capable of attending to different concepts such as subject of the scene, the activity represented, and the surrounding location, for both vision and text. We provide additional visualizations in \cref{sec:morevisualizations}.

\section{Discussion}
\label{sec:discussion}

Humans utilize a prior for compositionality to coherently represent different subsets of salient aspects of the world. We can enrich our comprehension by selectively attending to new aspects or specialize to a task by filtering out distracting ones. We introduce \Sparo with a goal of imparting a similar prior to transformers in representation learning frameworks by partitioning encodings into separately-attended concepts. Although we see evidence of disentanglement through positive results of post-hoc concept selection and visualization of attention maps, we do not impose any explicit independence or disentanglement constraints. In our representation learning frameworks, the pressure for learning distinct concepts arises from the need to explain the factors of variation through the training objective with only the features that can be accumulated by a single head of attention with embedded queries and limited dimensionality. However, auxiliary training objectives to impose stronger distributional conditions on the learned concepts can be explored as promising future directions. Beyond pretraining, we highlight that more sophisticated post-hoc concept selection approaches than explored in our work, like using human interaction or set cover algorithms, can further improve the downstream utility of \Sparo.

\section*{Acknowledgments}

This research was funded by Sony and enabled in part by compute resources provided by the Digital Research Alliance of Canada, Mila, and Sony.

{
    \bibliographystyle{splncs04}
    \bibliography{main}
}

\appendix
\counterwithin{figure}{section}
\counterwithin{table}{section}
\counterwithin{equation}{section}

\chapter*{Appendix}
\section{PyTorch implementation of \Sparo}
\label{sec:sparopytorch}

We provide an example implementation of \Sparo in PyTorch~\cite{paszke2019pytorch} as follows:
\begin{python}[basicstyle=\ttfamily\pythoncodesize]
class Sparo(nn.Module):
    def __init__(self, d, L, V, D, grp_size=1):
        """See Sec. 3 for descriptions of d, L, V, D.
        Each key/value projection weight is shared between grp_size slots."""
        super().__init__()
        self.L, self.V, self.D = L, V, D
        self.grp_size, self.nkeys = grp_size, L // grp_size
        self.scale = D ** -0.5

        self.q = nn.Parameter(torch.randn(1, self.nkeys, grp_size, 1, D))
        self.KVproj = nn.Linear(d, self.nkeys * D)
        self.Wproj = nn.Linear(D, V)

        nn.init.xavier_uniform_(self.KVproj.weight)
        nn.init.zeros_(self.KVproj.bias)
        nn.init.zeros_(self.Wproj.bias)

    def forward(self, x, eos_idx=None):
        """Provide eos_idx for text to prevent attending to positions after it."""
        batch_size, n, d = x.shape
        kv = self.KVproj(x).view(batch_size, n, self.nkeys, 1, self.D)
        kv = kv.expand(-1, -1, -1, self.grp_size, -1).permute(0, 2, 3, 1, 4)

        attn = (self.q * self.scale) @ kv.transpose(-2, -1)
        if eos_idx is not None:
            attn_mask = (
                torch.arange(n, device=eos_idx.device)[None, None, None, None, :]
                > eos_idx[:, None, None, None, None]
            )
            attn_mask = torch.where(
                attn_mask, -torch.inf, torch.zeros_like(attn_mask, dtype=attn.dtype)
            )
            attn += attn_mask
        attn = attn.softmax(dim=-1)
        x = (attn @ kv).view(batch_size, self.L, self.D)
        return self.Wproj(x)
\end{python}

Here, we optionally utilize multi-query attention~\cite{shazeer2019fast} to reduce the model size further by setting \pyth{grp_size>1}, allowing larger values of $L$ with the same resources. For example, when \pyth{grp_size=2} and \pyth{L=64}, only $32$ $\mK_l$ parameters are created.
We encourage practitioners to explore optimal \pyth{grp_size} settings for their applications.

\section{Additional results}
\label{sec:expadditional}

\subsection{Hyperparameters $L$ and $V$}
\label{sec:exp:lv}

\begin{figure}[tb]
    \centering
    \hfill\begin{subfigure}{0.475\linewidth}
        \centering
        \includegraphics[width=\linewidth]{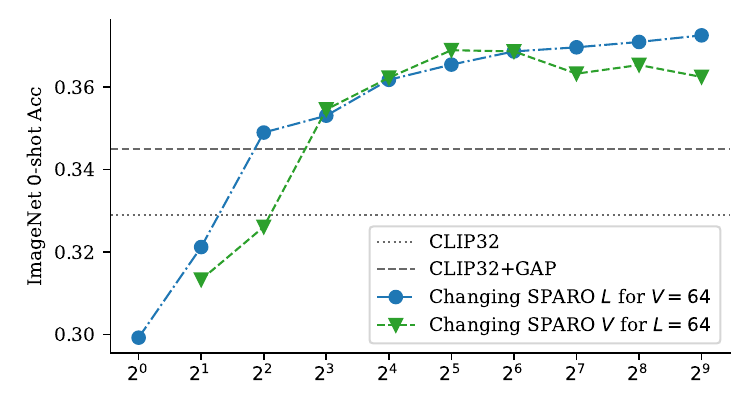}\\
        \vspace{-0.1em}
        \includegraphics[width=\linewidth]{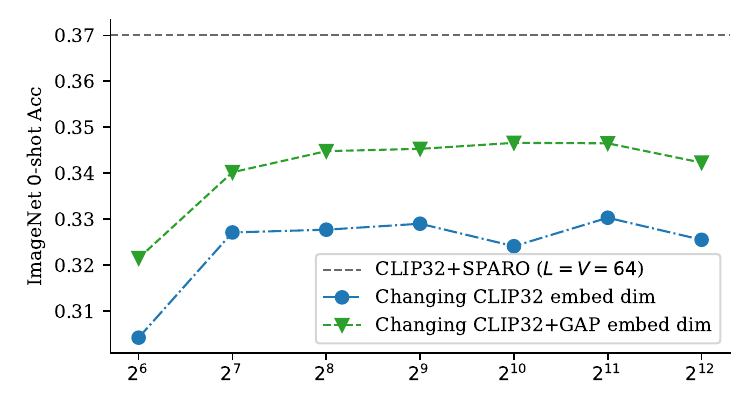}
        \caption{ImageNet zero-shot accuracy.}
    \end{subfigure}
    \hfill\begin{subfigure}{0.475\linewidth}
        \centering
        \includegraphics[width=\linewidth]{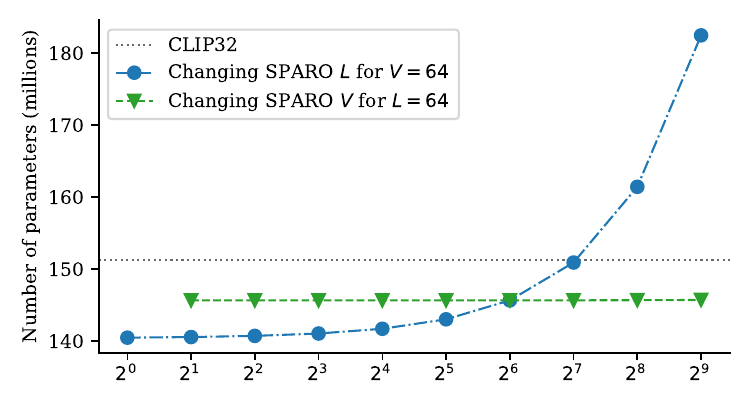}\\
        \vspace{-0.1em}
        \includegraphics[width=\linewidth]{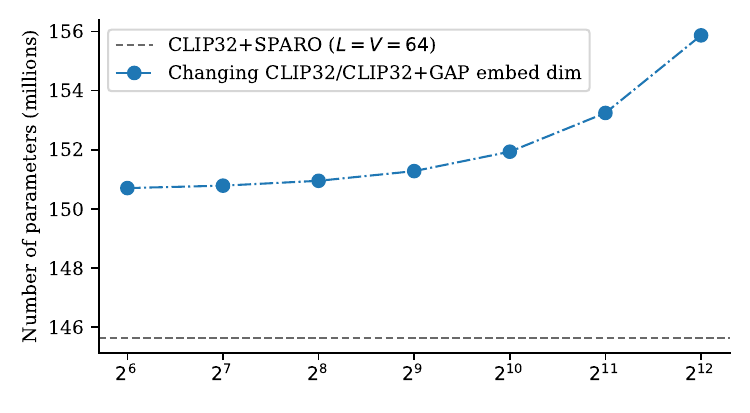}
        \caption{Number of model parameters.}
    \end{subfigure}
    \hfill\hfill%
    \caption{Effect of varying $L$ and $V$ values for CLIP$^{32}$+\Sparo, and the embedding size for CLIP$^{32}$ and CLIP$^{32}$+GAP, when training on CC15M.}
    \label{fig:lv}
\end{figure}

\Sparo encodings are concatenations of $L$ separately-attended slots of $V$ dimensions each. We study the effect of $L$ and $V$ with CLIP$^{32}$+\Sparo trained on CC15M, comparing its ImageNet zero-shot accuracy and model size with that of CLIP$^{32}$ and CLIP$^{32}$+GAP in \cref{fig:lv}. Generally, performance is poor for very low values of $L$ and $V$, and improves when they are increased. Improvements remain monotonic with $L$ in our experimental range, but the returns diminish for larger values of $L$. For each increment of $L$, we add additional $\vq_l$ and $\mK_l$ parameters to the model, and considering the trade-off between model size and performance gains becomes vital. Unlike $L$, increasing $V$ beyond optimal values leads to a performance drop. With very large values of $V$, we produce low-rank slots that can potentially encode arbitrary harmful biases for downstream tasks.

\subsection{Comparison with other bottlenecks}
\label{sec:otherbots}

\begin{table}[p!]
    \caption{Comparing zero-shot recognition, robustness, and SugarCrepe compositionality for CLIP models trained on Conceptual Captions with GAP, attentional pooling \cite{lee2019set,yu2022coca}, a linear (affine) information bottleneck, and a discrete bottleneck using Finite Discrete Tokens (FDT)~\cite{chen2023revisiting}.}
    \label{tab:bottlenecks}
    \centering
    \begin{tabular}{@{}llccccccc@{}}
        \toprule
        \multirow{2}[2]{*}{Train} &\multirow{2}[2]{*}{Model} &\multicolumn{5}{c}{ImageNet-} &\multirow{2}[2]{*}{\makecell{Object\\Net}} &\multirow{2}[2]{*}{\makecell{Sugar\\Crepe}} \\\cmidrule{3-7}
        & &V1 &V2 &Sketch &A &R & & \\
        \midrule
        \multirow{12}[2]{*}{CC3M} &CLIP$^{32}$ $^{(\mathcal{C})}$ &0.109 &0.091 &0.053 &0.025 &0.154 &0.064 &0.578 \\
        &$\mathcal{C}$+GAP &0.124 &0.105 &0.064 &0.025 &0.178 &0.078 &0.582 \\
        &$\mathcal{C}$+AttnPool &0.114 &0.092 &0.045 &0.026 &0.149 &0.070 &0.589 \\
        &$\mathcal{C}$+LinearBotneck ($64 \to 512$) &0.097 &0.083 &0.039 &0.025 &0.126 &0.060 &0.578 \\
        &$\mathcal{C}$+LinearBotneck ($512 \to 4096$) &0.108 &0.088 &0.047 &0.026 &0.139 &0.057 &0.591 \\
        &$\mathcal{C}$+DiscreteBotneck (FDT) &0.123 &0.101 &0.059 &\textbf{0.028} &0.166 &0.069 &0.583 \\
        &$\mathcal{C}$+\Sparo &\textbf{0.132} &\textbf{0.113} &\textbf{0.072} &0.027 &\textbf{0.190} &\textbf{0.083} &\textbf{0.596} \\
        \cmidrule{2-9}
        &CLIP$^{16}$ $^{(\mathcal{C})}$ &0.141 &0.122 &0.068 &0.033 &0.177 &0.080 &0.611 \\
        &$\mathcal{C}$+GAP &0.156 &0.134 &0.069 &0.033 &0.187 &0.081 &0.616 \\
        &$\mathcal{C}$+AttnPool &0.161 &0.138 &0.074 &0.036 &0.199 &0.090 &0.599 \\
        &$\mathcal{C}$+LinearBotneck ($64 \to 512$) &0.127 &0.110 &0.052 &0.028 &0.143 &0.070 &0.601 \\
        &$\mathcal{C}$+LinearBotneck ($512 \to 4096$) &0.138 &0.118 &0.058 &0.032 &0.162 &0.079 &0.590 \\
        &$\mathcal{C}$+DiscreteBotneck (FDT) &0.161 &0.137 &0.073 &\textbf{0.037} &0.195 &\textbf{0.099} &0.600 \\
        &$\mathcal{C}$+\Sparo &\textbf{0.170} &\textbf{0.140} &\textbf{0.088} &0.035 &\textbf{0.221} &0.098 &\textbf{0.625} \\
        \midrule
        \multirow{12}[2]{*}{CC12M} &CLIP$^{32}$ $^{(\mathcal{C})}$ &0.304 &0.257 &0.201 &0.060 &0.405 &0.160 &0.691 \\
        &$\mathcal{C}$+GAP &0.320 &0.272 &0.221 &0.063 &0.426 &0.178 &0.692 \\
        &$\mathcal{C}$+AttnPool &0.290 &0.240 &0.170 &0.046 &0.357 &0.165 &0.679 \\
        &$\mathcal{C}$+LinearBotneck ($64 \to 512$) &0.272 &0.232 &0.163 &0.052 &0.338 &0.149 &0.675 \\
        &$\mathcal{C}$+LinearBotneck ($512 \to 4096$) &0.285 &0.238 &0.178 &0.052 &0.353 &0.151 &0.683 \\
        &$\mathcal{C}$+DiscreteBotneck (FDT) &0.307 &0.264 &0.206 &0.063 &0.424 &0.180 &\textbf{0.693} \\
        &$\mathcal{C}$+\Sparo &\textbf{0.344} &\textbf{0.293} &\textbf{0.251} &\textbf{0.069} &\textbf{0.468} &\textbf{0.200} &\textbf{0.693} \\
        \cmidrule{2-9}
        &CLIP$^{16}$ $^{(\mathcal{C})}$ &0.361 &0.311 &0.249 &0.091 &0.467 &0.218 &0.697 \\
        &$\mathcal{C}$+GAP &0.382 &0.330 &0.262 &0.101 &0.501 &0.241 &0.695 \\
        &$\mathcal{C}$+AttnPool &0.357 &0.305 &0.226 &0.085 &0.428 &0.226 &0.685 \\
        &$\mathcal{C}$+LinearBotneck ($64 \to 512$) &0.331 &0.286 &0.198 &0.073 &0.387 &0.187 &0.674 \\
        &$\mathcal{C}$+LinearBotneck ($512 \to 4096$) &0.344 &0.292 &0.226 &0.081 &0.428 &0.212 &0.688 \\
        &$\mathcal{C}$+DiscreteBotneck (FDT) &0.367 &0.313 &0.260 &0.086 &0.490 &0.238 &0.719 \\
        &$\mathcal{C}$+\Sparo &\textbf{0.406} &\textbf{0.350} &\textbf{0.298} &\textbf{0.113} &\textbf{0.559} &\textbf{0.268} &\textbf{0.723} \\
        \midrule
        \multirow{12}[2]{*}{CC15M} &CLIP$^{32}$ $^{(\mathcal{C})}$ &0.329 &0.279 &0.232 &0.070 &0.435 &0.178 &0.687 \\
        &$\mathcal{C}$+GAP &0.345 &0.294 &0.242 &0.077 &0.460 &0.192 &0.678 \\
        &$\mathcal{C}$+AttnPool &0.312 &0.262 &0.190 &0.065 &0.374 &0.168 &0.671 \\
        &$\mathcal{C}$+LinearBotneck ($64 \to 512$) &0.300 &0.251 &0.187 &0.054 &0.360 &0.159 &0.665 \\
        &$\mathcal{C}$+LinearBotneck ($512 \to 4096$) &0.313 &0.265 &0.203 &0.064 &0.383 &0.165 &0.663 \\
        &$\mathcal{C}$+DiscreteBotneck (FDT) &0.333 &0.282 &0.233 &0.071 &0.456 &0.197 &0.685 \\
        &$\mathcal{C}$+\Sparo &\textbf{0.370} &\textbf{0.313} &\textbf{0.271} &\textbf{0.083} &\textbf{0.510} &\textbf{0.213} &\textbf{0.710} \\
        \cmidrule{2-9}
        &CLIP$^{16}$ $^{(\mathcal{C})}$ &0.384 &0.337 &0.268 &0.105 &0.503 &0.238 &0.699 \\
        &$\mathcal{C}$+GAP &0.399 &0.343 &0.287 &0.114 &0.531 &0.252 &0.701 \\
        &$\mathcal{C}$+AttnPool &0.387 &0.329 &0.248 &0.100 &0.469 &0.233 &0.685 \\
        &$\mathcal{C}$+LinearBotneck ($64 \to 512$) &0.357 &0.306 &0.226 &0.085 &0.422 &0.209 &0.672 \\
        &$\mathcal{C}$+LinearBotneck ($512 \to 4096$) &0.372 &0.321 &0.245 &0.094 &0.455 &0.218 &0.683 \\
        &$\mathcal{C}$+DiscreteBotneck (FDT) &0.394 &0.338 &0.279 &0.124 &0.531 &0.254 &0.719 \\
        &$\mathcal{C}$+\Sparo &\textbf{0.437} &\textbf{0.378} &\textbf{0.317} &\textbf{0.145} &\textbf{0.579} &\textbf{0.279} &\textbf{0.730} \\
        \bottomrule
    \end{tabular}
\end{table}

\Sparo imposes a representational bottleneck by constraining each slot to the output of a single head of attention with fixed queries and limited dimensionality. In this section, we compare our bottleneck with two other representational bottlenecks: a linear (affine) information bottleneck and a discrete bottleneck using Finite Discrete Tokens (FDT)~\cite{chen2023revisiting}.

\paragraph{Setup.} We add the considered bottlenecks and \Sparo to CLIP$^{32}$ and CLIP$^{16}$, and train on different sizes of Conceptual Captions. We empirically verified that removing a block of the transformer backbone reduces the performance for all models. Still, we follow the typical setup of replacing the last transformer block of the backbone when using \Sparo. However, we do not handicap other bottleneck baselines, adding the corresponding bottlenecks after the last transformer block.

\paragraph{Linear information bottleneck details.} We implement a linear information bottleneck by setting the embedding size of standard CLIP to $m$ and adding an affine transformation to an output size of $M$ on top of it, with $m < M$. We evaluate two settings for the $m \to M$ bottleneck: $64 \to 512$ and $512 \to 4096$.

\paragraph{Discrete bottleneck details.} We use Finite Discrete Tokens (FDT)~\cite{chen2023revisiting} to implement a discrete bottleneck. FDT encodes images and texts as discrete codes belonging to a shared learned codebook. During training, a convex combination of the discrete codes is produced using the Sparsemax operation \cite{martins2016softmax}, which the authors show performs better than Softmax. We follow the hyperparameters used by the authors: the codebook contains 16,384 codes of 512 dimensions each.

\paragraph{Results.} We present our results in \cref{tab:bottlenecks}. First, we see that a simple linear bottleneck performs worse than standard CLIP in a majority of settings, indicating that the mechanism behind the representational bottleneck is important. As expected, the discrete bottleneck widely outperforms or stays comparable to standard CLIP. However, note that global average pooling (GAP) continues to shine as a strong baseline, outperforming the discrete bottleneck in most settings. Finally, we verify that \Sparo outperforms or remains comparable to the best baselines across all settings.

\subsection{CLIP with ResNet encoders}
\label{sec:resnets}

\begin{table}[tb]
    \caption{Comparing zero-shot recognition, robustness, and SugarCrepe compositionality for CLIP models trained on CC15M with ResNet-50 encoders.}
    \label{tab:resnets}
    \centering
    \resizebox{\linewidth}{!}{%
    \begin{tabular}{@{}ccccccccc@{}}
        \toprule
        \multirow{2}[2]{*}{Model} &\multirow{2}[2]{*}{Params} &\multicolumn{5}{c}{ImageNet-} &\multirow{2}[2]{*}{\makecell{Object\\Net}} &\multirow{2}[2]{*}{\makecell{Sugar\\Crepe}} \\\cmidrule{3-7}
        & &V1 &V2 &Sketch &A &R & & \\
        \midrule
        CLIP$^\text{R50}$ ($M=1024$) & 102M &0.298 &0.272 &0.227 &0.090 &0.430 &0.220 &0.679 \\
        CLIP$^\text{R50}$ ($M=4096$) &110M &0.295 &0.275 &0.226 &0.086 &0.422 &0.213 &0.683 \\
        CLIP$^\text{R50}$+\Sparo ($L=128, V=32, D=32$) &97M &0.338 &0.297 &\textbf{0.263} &\textbf{0.101} &0.476 &\textbf{0.256} &0.713 \\
        CLIP$^\text{R50}$+\Sparo ($L=128, V=32, D=64$) &108M &\textbf{0.344} &\textbf{0.304} &\textbf{0.263} &0.093 &\textbf{0.477} &0.252 &\textbf{0.719} \\
        \bottomrule
    \end{tabular}}
\end{table}

We compare CLIP models with residual network (ResNet) \cite{he2016deep} encoders, trained on CC15M with and without \Sparo, in \cref{tab:resnets}. Our findings are consistent with those observed with ViT encoders, demonstrating that \Sparo improves model performance across SugarCrepe, ImageNet, and robustness benchmarks without increasing model size.

\subsection{Fine-grained SugarCrepe performance}
\label{sec:finesugarcrepe}

\begin{table}[tb]
  \caption{Fine-grained SugarCrepe zero-shot accuracies for CLIP models trained on Conceptual Captions and LAION-400M.}
  \label{tab:sugarcrepe}
  \centering
  \begin{tabular}{@{}llcccp{0.17em}ccp{0.17em}ccc@{}}
      \toprule
      \multirow{2}[2]{*}{Train} &\multirow{2}[2]{*}{Model} &\multicolumn{3}{c}{Replace} & &\multicolumn{2}{c}{Swap} & &\multicolumn{2}{c}{Add} &\multirow{2}[2]{*}{Avg} \\\cmidrule{3-5}\cmidrule{7-8}\cmidrule{10-11}
      & &Object &Attribute &Relation & &Object &Attribute & &Object &Attribute & \\
      \midrule
      \multirow{3}{*}{CC3M} &CLIP$^{16}$ $^{(\mathcal{C})}$ &0.741 &0.673 &0.575 & &0.524 &0.584 & &\textbf{0.615} &0.566 &0.611 \\
      &$\mathcal{C}$+GAP &0.725 &0.675 &0.575 & &0.524 &\textbf{0.598} & &0.611 &\textbf{0.601} &0.616 \\
      &$\mathcal{C}$+\Sparo &\textbf{0.763} &\textbf{0.689} &\textbf{0.585} & &\textbf{0.541} &0.592 & &0.611 &0.592 &\textbf{0.625} \\
      \midrule
      \multirow{3}{*}{CC12M} &CLIP$^{16}$ $^{(\mathcal{C})}$ &0.860 &0.737 &\textbf{0.666} & &0.593 &0.619 & &0.743 &0.659 &0.697 \\
      &$\mathcal{C}$+GAP &0.873 &0.747 &0.647 & &0.573 &0.619 & &0.741 &\textbf{0.666} &0.695 \\
      &$\mathcal{C}$+\Sparo &\textbf{0.881} &\textbf{0.784} &0.649 &  &\textbf{0.642} &\textbf{0.674} & &\textbf{0.768} &0.659 &\textbf{0.723} \\
      \midrule
      \multirow{3}{*}{CC15M} &CLIP$^{16}$ $^{(\mathcal{C})}$ &0.866 &0.753 &0.664 & &0.577 &0.613 & &0.761 &0.660 &0.699 \\
      &$\mathcal{C}$+GAP &0.875 &0.770 &0.649 & &0.541 &0.640 & &0.758 &0.672 &0.701 \\
      &$\mathcal{C}$+\Sparo &\textbf{0.896} &\textbf{0.798} &\textbf{0.666} & &\textbf{0.593} &\textbf{0.700} & &\textbf{0.779} &\textbf{0.679} &\textbf{0.730} \\
      \midrule
      \multirow{3}{*}{L400M} &CLIP$^{32}$ $^{(\mathcal{C})}$ &0.921 &\textbf{0.826} &\textbf{0.678} & &\textbf{0.605} &0.659 & &0.815 &0.735 &0.748 \\
      &$\mathcal{C}$+GAP &0.920 &0.797 &0.661 & &0.565 &0.631 & &0.815 &0.733 &0.732 \\
      &$\mathcal{C}$+\Sparo &\textbf{0.927} &0.822 &\textbf{0.678} & &0.585 &\textbf{0.719} & &\textbf{0.866} &\textbf{0.794} &\textbf{0.770} \\
      \bottomrule
  \end{tabular}
\end{table}

The SugarCrepe~\cite{hsieh2023sugarcrepe} benchmark comprises of seven compositional manipulations of captions for a subset of MS~COCO \cite{lin2014microsoft}. The benchmark evaluates the zero-shot accuracy of vision-language models for picking the ground-truth captions over the generated hard negatives. We provide the fine-grained results of our models for each of the seven SugarCrepe manipulations in \cref{tab:sugarcrepe}. Individual categories often demonstrate trade-offs amongst each other, but the averaged performance reveals consistent trends of \Sparo outperforming the baselines. Additionally, we show that the compositionality exposed in learned \Sparo encodings across different categories can be enhanced by performing post-hoc concept selection, demonstrated in \cref{sec:exp:naiveslots}.

\subsection{Winoground performance}
\label{sec:winoground}

\begin{table}[tb]
  \caption{Winoground zero-shot accuracies for CLIP models trained on Conceptual Captions and LAION-400M.}
  \label{tab:winoground}
  \centering
  \begin{tabular}{@{}llccc@{}}
      \toprule
      Train &Model &Text &Image &Group \\
      \midrule
      \multirow{3}{*}{CC3M} &CLIP$^{16}$ $^{(\mathcal{C})}$ &0.205 &0.075 &0.052 \\
      &$\mathcal{C}$+GAP &0.190 &0.097 &0.065 \\
      &$\mathcal{C}$+\Sparo &\textbf{0.215} &\textbf{0.123} &\textbf{0.072} \\
      \midrule
      \multirow{3}{*}{CC12M} &CLIP$^{16}$ $^{(\mathcal{C})}$ &\textbf{0.248} &0.087 &0.052 \\
      &$\mathcal{C}$+GAP &0.245 &0.095 &0.052 \\
      &$\mathcal{C}$+\Sparo &0.245 &\textbf{0.102} &\textbf{0.058} \\
      \midrule
      \multirow{3}{*}{CC15M} &CLIP$^{16}$ $^{(\mathcal{C})}$ &0.243 &0.085 &0.052 \\
      &$\mathcal{C}$+GAP &0.245 &0.078 &0.052 \\
      &$\mathcal{C}$+\Sparo &\textbf{0.273} &\textbf{0.102} &\textbf{0.075} \\
      \midrule
      \multirow{3}{*}{L400M} &CLIP$^{32}$ $^{(\mathcal{C})}$ &\textbf{0.287} &0.110 &0.083 \\
      &$\mathcal{C}$+GAP &0.265 &0.132 &0.087 \\
      &$\mathcal{C}$+\Sparo &0.278 &\textbf{0.140} &\textbf{0.097} \\
      \bottomrule
  \end{tabular}
\end{table}

We evaluate CLIP+\Sparo against CLIP and CLIP+GAP on Winoground~\cite{thrush2022winoground}, and report the metrics suggested by the benchmark. Given image-caption pairs $(I_0, C_0)$ and $(I_1, C_1)$, the `Text' score averages whether the model can select the correct captions $C_0$ and $C_1$ given the images $I_0$ and $I_1$ respectively. The `Image' score tallies whether the model can select the correct images $I_0$ and $I_1$ for the captions $C_0$ and $C_1$ respectively. The `Group' score combines these tests, and evaluates whether the model can select both the correct image given the caption, and the correct caption given the image, for combinations of image-text pairs. We report our results in \cref{tab:winoground}, showing superior performance of \Sparo in all of the `Image' and `Group' evaluations. For `Text', we find that \Sparo performs as well as or better than GAP, but CLIP without GAP outperforms both in two out of our four training setups.

\subsection{Absolute VTAB results}
\label{sec:absvtab}

\begin{table}[tb]
    \caption{Absolute VTAB benchmark results corresponding to the relative differences illustrated in \cref{fig:vtab}.}
    \label{tab:absvtab}
    \centering
    \begin{tabular}{@{}lccp{0.17em}ccp{0.17em}ccp{0.17em}cc@{}}
        \toprule
\multirow{2}{*}{VTAB dataset} &\multicolumn{2}{c}{CC3M ($\mathcal{C}^{16}$)} & &\multicolumn{2}{c}{CC12M ($\mathcal{C}^{16}$)} & &\multicolumn{2}{c}{CC15M ($\mathcal{C}^{16}$)} & &\multicolumn{2}{c}{L400M ($\mathcal{C}^{32}$)} \\\cmidrule{2-3}\cmidrule{5-6}\cmidrule{8-9}\cmidrule{11-12}
 &GAP &\Sparo & &GAP &\Sparo & &GAP &\Sparo & &GAP &\Sparo \\
\midrule
\texttt{cifar100} &0.163 &\textbf{0.168} & &0.330 &\textbf{0.390} & &0.397 &\textbf{0.449} & &0.707 &\textbf{0.742} \\
\texttt{resisc45} &0.169 &\textbf{0.184} & &0.326 &\textbf{0.359} & &0.357 &\textbf{0.360} & &0.527 &\textbf{0.563} \\
\texttt{eurosat} &0.193 &\textbf{0.226} & &0.276 &\textbf{0.335} & &0.246 &\textbf{0.292} & &0.480 &\textbf{0.513} \\
\texttt{dtd} &0.104 &\textbf{0.121} & &0.162 &\textbf{0.234} & &0.181 &\textbf{0.195} & &0.502 &\textbf{0.506} \\
\texttt{svhn} &0.074 &\textbf{0.097} & &0.087 &\textbf{0.098} & &0.084 &\textbf{0.125} & &0.236 &\textbf{0.458} \\
\texttt{caltech101} &0.539 &\textbf{0.562} & &0.731 &\textbf{0.744} & &\textbf{0.752} &0.748 & &0.828 &\textbf{0.829} \\
\texttt{smallnorb\_elevation} &0.114 &\textbf{0.119} & &0.100 &\textbf{0.108} & &\textbf{0.138} &0.116 & &0.102 &\textbf{0.108} \\
\texttt{smallnorb\_azimuth} &\textbf{0.055} &0.053 & &0.052 &\textbf{0.053} & &\textbf{0.058} &0.057 & &0.049 &\textbf{0.053} \\
\texttt{clevr\_closest\_obj} &\textbf{0.224} &0.194 & &0.151 &\textbf{0.197} & &\textbf{0.109} &0.095 & &\textbf{0.204} &0.082 \\
\texttt{clevr\_count} &0.105 &\textbf{0.132} & &0.227 &\textbf{0.232} & &0.136 &\textbf{0.151} & &\textbf{0.189} &0.164 \\
\texttt{pcam} &0.487 &\textbf{0.495} & &0.477 &\textbf{0.535} & &0.502 &\textbf{0.572} & &\textbf{0.549} &0.495 \\
\texttt{sun397} &0.261 &\textbf{0.263} & &0.464 &\textbf{0.489} & &0.505 &\textbf{0.519} & &\textbf{0.671} &0.663 \\
\texttt{dmlab} &0.148 &\textbf{0.172} & &\textbf{0.172} &0.135 & &0.134 &\textbf{0.171} & &\textbf{0.202} &0.141 \\
\texttt{kitti\_closest\_dist} &0.118 &\textbf{0.322} & &\textbf{0.421} &0.392 & &0.249 &\textbf{0.429} & &\textbf{0.245} &0.125 \\
\texttt{diabetic\_ret} &0.066 &\textbf{0.147} & &\textbf{0.165} &0.118 & &0.166 &\textbf{0.174} & &\textbf{0.047} &0.023 \\
\texttt{dsprites\_xpos} &\textbf{0.033} &0.028 & &\textbf{0.030} &0.029 & &0.035 &\textbf{0.044} & &\textbf{0.032} &0.030 \\
\texttt{dsprites\_orientation} &\textbf{0.027} &0.020 & &\textbf{0.029} &0.026 & &\textbf{0.019} &0.015 & &0.017 &\textbf{0.025} \\
\texttt{pets} &0.099 &\textbf{0.104} & &\textbf{0.587} &0.575 & &\textbf{0.595} &0.579 & &0.860 &\textbf{0.869} \\
\texttt{flowers} &0.119 &\textbf{0.131} & &\textbf{0.298} &0.287 & &\textbf{0.324} &0.311 & &0.659 &\textbf{0.673} \\
        \bottomrule
    \end{tabular}
\end{table}

In \cref{tab:absvtab}, we present the absolute numbers behind the relative improvements of \Sparo illustrated in \cref{fig:vtab} for the VTAB benchmark \cite{zhai2019large}.

\subsection{Additional visualizations}
\label{sec:morevisualizations}

\begin{figure}[tb]
    \visspacing
    \centering
    \begin{subfigure}{0.24\linewidth}
        \centering
        \includegraphics[width=\linewidth,height=\visheight,keepaspectratio]{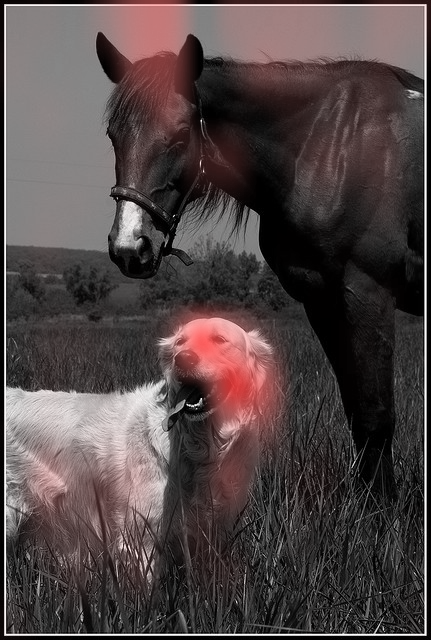}
        \caption*{\sffamily\viscaptionsize A brown \textcolor{red}{horse} stands over a light colored \textcolor{red}{\textbf{dog}}}
    \end{subfigure}
    \begin{subfigure}{0.24\linewidth}
        \centering
        \includegraphics[width=\linewidth,height=\visheight,keepaspectratio]{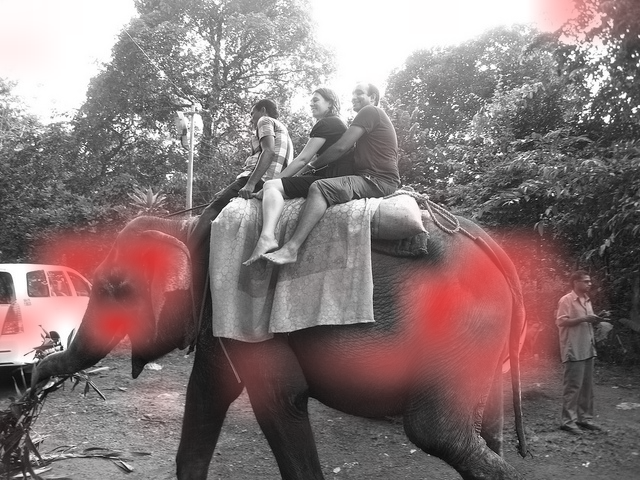}
        \caption*{\sffamily\viscaptionsize Three people ride on the back of an \textcolor{red}{\textbf{elephant}}}
    \end{subfigure}
    \begin{subfigure}{0.24\linewidth}
        \centering
        \includegraphics[width=\linewidth,height=\visheight,keepaspectratio]{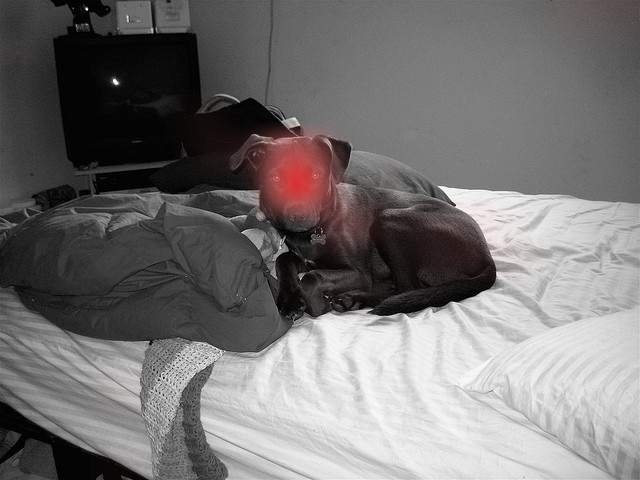}
        \caption*{\sffamily\viscaptionsize \textcolor{red}{\textbf{Dog}} lies on bed looking towards the camera}
    \end{subfigure}
    \begin{subfigure}{0.24\linewidth}
        \centering
        \includegraphics[width=\linewidth,height=\visheight,keepaspectratio]{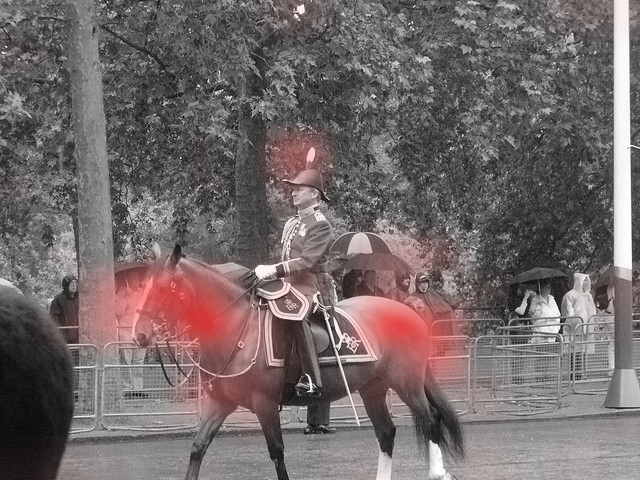}
        \caption*{\sffamily\viscaptionsize A man in uniform rides a \textcolor{red}{\textbf{horse}} down the street}
    \end{subfigure}\\
    \begin{subfigure}{0.24\linewidth}
        \centering
        \includegraphics[width=\linewidth,height=\visheight,keepaspectratio]{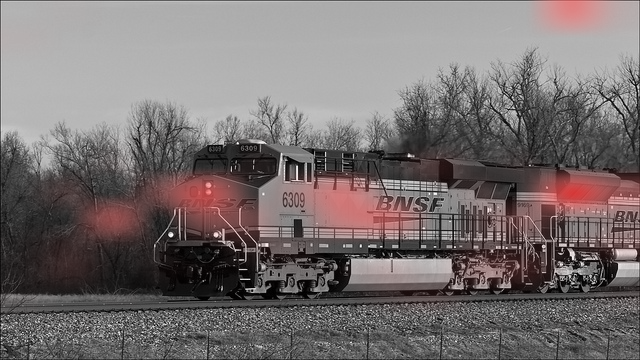}
        \caption*{\sffamily\viscaptionsize The \textcolor{red}{train \textbf{engine}} number 6309 is operated by BNSF}
    \end{subfigure}
    \begin{subfigure}{0.24\linewidth}
        \centering
        \includegraphics[width=\linewidth,height=\visheight,keepaspectratio]{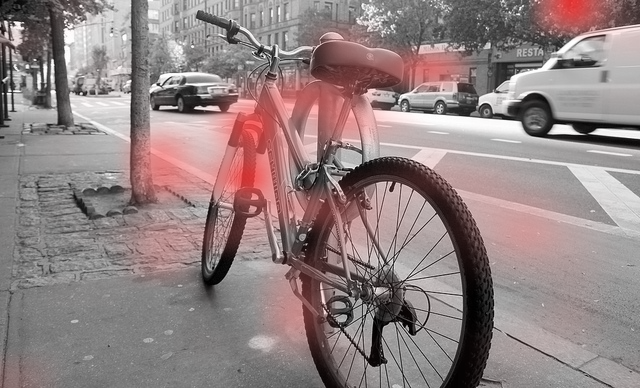}
        \caption*{\sffamily\viscaptionsize A blue \textcolor{red}{\textbf{bicycle}} sits on a sidewalk near a street}
    \end{subfigure}
    \begin{subfigure}{0.24\linewidth}
        \centering
        \includegraphics[width=\linewidth,height=\visheight,keepaspectratio]{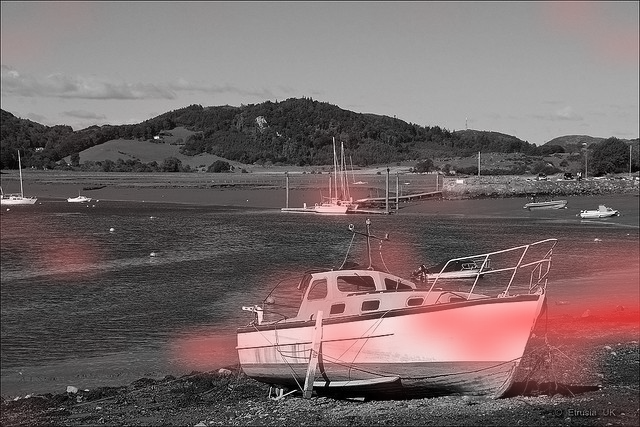}
        \caption*{\sffamily\viscaptionsize A \textcolor{red}{\textbf{boat}} stowed up on the beaches on sand}
    \end{subfigure}
    \begin{subfigure}{0.24\linewidth}
        \centering
        \includegraphics[width=\linewidth,height=\visheight,keepaspectratio]{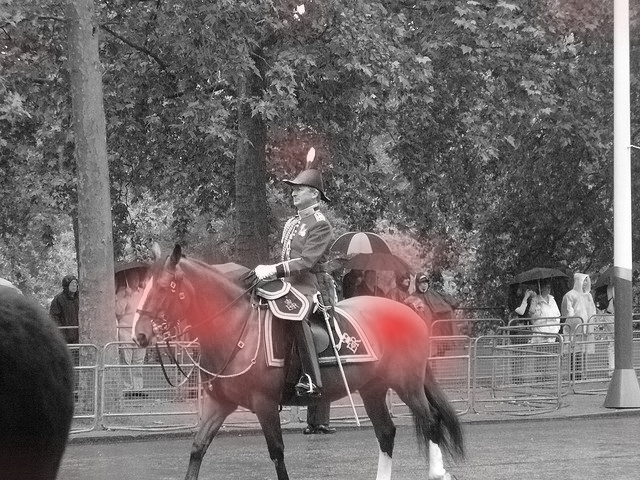}
        \caption*{\sffamily\viscaptionsize A man in uniform rides a \textcolor{red}{\textbf{horse}} down the street}
    \end{subfigure}\\
    \begin{subfigure}{0.24\linewidth}
        \centering
        \includegraphics[width=\linewidth,height=\visheight,keepaspectratio]{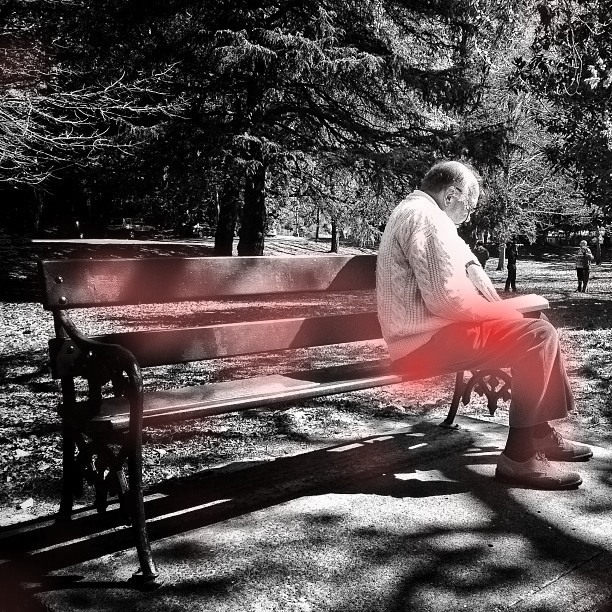}
        \caption*{\sffamily\viscaptionsize A man \textcolor{red}{sitting} alone on a park \textcolor{red}{\textbf{bench}} in a park}
    \end{subfigure}
    \begin{subfigure}{0.24\linewidth}
        \centering
        \includegraphics[width=\linewidth,height=\visheight,keepaspectratio]{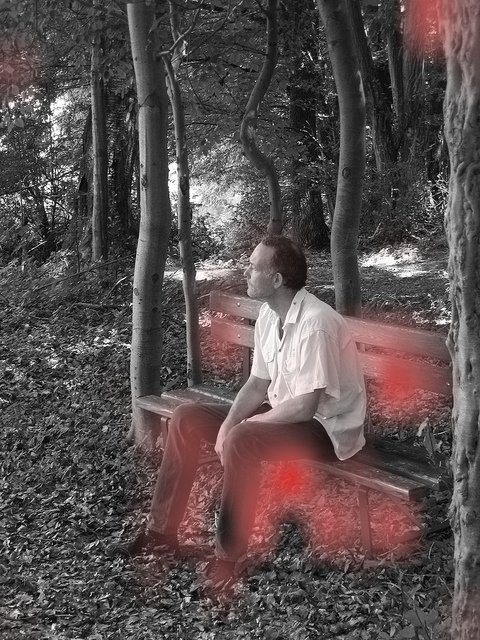}
        \caption*{\sffamily\viscaptionsize A man \textcolor{red}{sits} on a \textcolor{red}{\textbf{bench}} underneath trees}
    \end{subfigure}
    \begin{subfigure}{0.24\linewidth}
        \centering
        \includegraphics[width=\linewidth,height=\visheight,keepaspectratio]{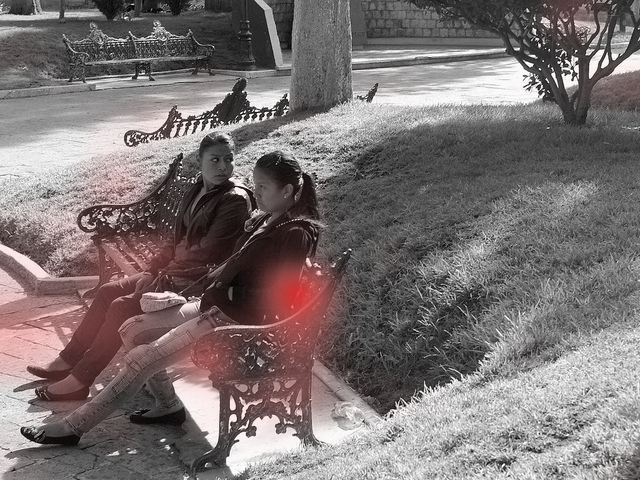}
        \caption*{\sffamily\viscaptionsize Two women \textcolor{red}{sit} next to each other on a park \textcolor{red}{\textbf{bench}}}
    \end{subfigure}
    \begin{subfigure}{0.24\linewidth}
        \centering
        \includegraphics[width=\linewidth,height=\visheight,keepaspectratio]{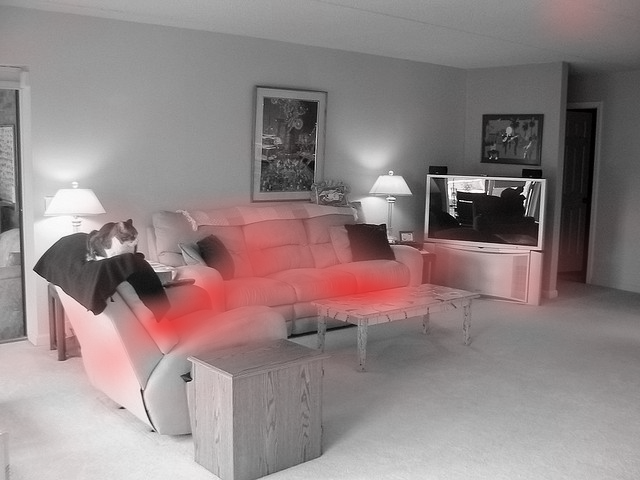}
        \caption*{\sffamily\viscaptionsize A room with a \textcolor{red}{\textbf{couch}}, \textcolor{BrickRed}{chair}, table and television in it}
    \end{subfigure}
    \caption{Additional visualizations of \textcolor{red}{attended} image and text positions for three \Sparo slots (one per row). We surmise that the \Sparo concepts from top to bottom represent animals, transportation, and seats. Notice that top-right and center-right examples have similar attention masks over `horse,' but consider different aspects of the concept -- one as an animal, another as a mode of transportation.}
    \label{fig:morevisualization}
\end{figure}

We present additional examples of learned \Sparo concepts in \cref{fig:morevisualization} following the filtering described in \cref{sec:exp:visualizations}. Each row illustrates the attended positions of the single-head attention operation for the same \Sparo slot across examples from MS~COCO~\cite{lin2014microsoft}. We reuse the same last input for the first two visualized slots to highlight cases where visually similar attention maps can correspond to distinct concepts. Looking at attended positions allows us to interpret \emph{where} each slot takes information from, but not \emph{how} that information is used.

\section{Reproducibility}
\label{sec:hyperparameters}

\begin{table}[tb]
    \caption{Training datasets and the train split sizes used in our experiments. For CC3M, CC12M, and LAION-400M, our downloaded dataset sizes are smaller than the originally published sizes due to images becoming unavailable for download over time.}
    \label{tab:datasets}
    \centering
    \begin{tabular}{@{}lccccc@{}}
        \toprule
        Training dataset &\# Classes &Published size &Downloaded size &Download rate \\
        \midrule
        CC3M \cite{sharma2018conceptual} &\multirow{4}{*}{N/A} &3,318,333 &2,795,293 &84.24\% \\
        CC12M \cite{changpinyo2021cc12m} & &12,423,374 &10,030,127 &80.74\% \\
        CC15M (CC3M+CC12M) & &N/A &12,825,420 &N/A \\
        LAION-400M \cite{schuhmann2021laion} & &413,871,335 &319,707,672 &77.25\% \\
        \midrule
        CIFAR-10 \cite{krizhevsky2009learning} &10 &\multicolumn{2}{c}{50,000} &100\% \\
        CIFAR-100 \cite{krizhevsky2009learning} &100 &\multicolumn{2}{c}{50,000} &100\% \\
        ImageNet \cite{deng2009imagenet} &1000 &\multicolumn{2}{c}{1,281,167} &100\% \\
        \bottomrule
    \end{tabular}
\end{table}

\begin{table}[tb]
    \caption{Open source repositories used and their modifications for our experiments.}
    \label{tab:repositories}
    \centering
    \begin{tabular}{@{}lcc@{}}
        \toprule
        Task &Open-source repo &Summary of changes \\
        \midrule
        CLIP Training &\multirow{2}{*}{OpenCLIP \cite{openclip}} &\multirow{2}{*}{Added \Sparo and text AttPool} \\
        CLIP ImageNet $0$-shot & & \\\cmidrule{2-3}
        CLIP SugarCrepe $0$-shot &SugarCrepe \cite{sugarcreperepo} &Support \Sparo slot selection \\\cmidrule{2-3}
        CLIP linear probe &\multirow{3}{*}{CLIP benchmark \cite{clipbenchmark}} &\multirow{3}{*}{} \\
        CLIP $0$-shot robustness & & \\
        CLIP $0$-shot retrieval & & \\
        \midrule
        DINO Training &\multirow{3}{*}{DINO \cite{dinorepo}} &\multirow{3}{*}{Added \Sparo} \\
        DINO $k$-NN & & \\
        DINO linear probe & & \\
        \bottomrule
    \end{tabular}
\end{table}

We present the details of the training datasets used in our experiments in \cref{tab:datasets}. Note that the datasets we use for training our CLIP models are potentially smaller than those used by prior work, as images become unavailable for download over time for the datasets we consider. \cref{tab:repositories} lists the open-source repositories we utilized and modified for training and for the evaluations presented in our work. Finally, note that \cref{sec:sparopytorch} provides example PyTorch code for implementing the \Sparo module.

\subsection{CLIP}
\label{sec:cliphyperparams}

\begin{table}[tb]
    \caption{Hyperparameters for CLIP training using OpenCLIP~\cite{openclip}.}
    \label{tab:cliptrainhyper}
    \centering
    \begin{tabular}{@{}lccccc@{}}
        \toprule
        \multirow{2}[2]{*}{Hyperparameter} & &\multicolumn{4}{c}{Hyperparameter value} \\\cmidrule{2-6}
        & Dataset:&CC3M &CC12M &CC15M &L400M \\\midrule
        \multicolumn{2}{l}{Epochs} &32 &25 &25 &32 \\
        \multicolumn{2}{l}{Global batch size} &\multicolumn{3}{c}{4096} &32768 \\
        \multicolumn{2}{l}{Warmup steps} &\multicolumn{3}{c}{3600} &2000 \\
        \multicolumn{2}{l}{Weight decay} &\multicolumn{3}{c}{0.5} &0.2 \\
        \multicolumn{2}{l}{Precision} &\multicolumn{4}{c}{Automatic mixed precision with BFloat16} \\
        \bottomrule
    \end{tabular}
\end{table}

\paragraph{CLIP training.} We use the open-source OpenCLIP~\cite{openclip} project for training our CLIP models. The important hyperparameters that we change from the OpenCLIP defaults for training our models are listed in \cref{tab:cliptrainhyper}.

\paragraph{Linear probe training.} We use the open-source CLIP benchmark \cite{clipbenchmark} project for training and evaluating linear probes for our trained OpenCLIP checkpoints. We train our linear probes on $\ell_2$-normalized CLIP encodings using the AdamW \cite{loshchilov2017decoupled} optimizer with a learning rate of $0.1$ for 10 epochs. For each setting, we first sweep for the optimal weight decay hyperparameter by training on a subset of the training split and evaluating on the remaining samples. Finally, we train on the full training set with the chosen weight decay value. The design of the hyperparameter sweep follows that of \cite{radford2021learning}.

\paragraph{Training masks on SugarCrepe.} \label{sec:maskingdetails}
Our training set is prepared by combining the splits of the SugarCrepe benchmark without any extra balancing. For each training sample, we obtain three encodings: the image encoding $\vy^\text{i}$, positive text encoding $\vy^\text{p}$, and the negative text encoding $\vy^\text{n}$. We produce the masked image encoding $\vy^\text{i} = \vm \odot \vy$, where $\vm$ is generated as $\vm = \sigma\left(\frac{1}{4}\max(100, \exp(\alpha))\,\vtheta\right)$. Here, $\sigma$ is the element-wise sigmoid function, $\alpha \in \R$ and $\vtheta \in \R^{M'}$ are learned parameters, and $\max(100, \exp(\alpha))/4$ is the (inverse) temperature for the sigmoid that we found to help in faster convergence and stability of training. We have $M'=L$ for per-slot masking and $M'=M$ for per-dimension masking. We use the cosine similarities $\vy^\text{i} \cdot \vy^\text{p} / \norm{\vy^\text{i}}\norm{\vy^\text{p}}$ and $\vy^\text{i} \cdot \vy^\text{n} / \norm{\vy^\text{i}}\norm{\vy^\text{n}}$, scaled by $\max(100, \exp(\alpha))$ similar to CLIP~\cite{radford2021learning}, as the logits for the cross-entropy loss that we minimize. For each setting, we train for 100 epochs using SGD, with learning rate 0.02 and momentum 0.9. All settings converge before 100 epochs, and we pick the mask which achieves the best fit on standard SugarCrepe evaluation during training.

\subsection{DINO}
\label{sec:dinohyperparams}

We use the official DINO repository \cite{dinorepo} to train and evaluate our DINO models. For training, we follow the instructions for ``Vanilla DINO training,'' using a ViT-Small backbone trained on a single node with 8 GPUs for 100 epochs. Similarly, we use the default hyperparameters for $k$-nearest neighbors and linear probe evaluations. The linear probe is trained on the layer normalized \texttt{CLS} token outputs from the last 4 ViT blocks. When training the linear probe on DINO+\Sparo, we use the unnormalized \Sparo encoding instead of the last block's normalized \texttt{CLS} output.

\end{document}